%% file: main.tex
\documentclass[11pt]{article}

\usepackage[final]{acl}

\usepackage{times}
\usepackage{latexsym}

\usepackage[T1]{fontenc}

\usepackage[utf8]{inputenc}

\usepackage{microtype}

\usepackage{inconsolata}

\usepackage{graphicx}

\usepackage{subcaption} 
\usepackage{marginnote} 

\usepackage{comment} 
\usepackage{booktabs} 
\usepackage{xcolor} 
\usepackage{parskip} 
\usepackage{float} 
\usepackage{multirow} 
\usepackage{amsmath} 
\usepackage{amsfonts}
\usepackage{amssymb}
\usepackage{glossaries}
\usepackage{adjustbox}
\usepackage{array}
\usepackage{arydshln}
\usepackage{tabularx} 
\usepackage{siunitx}
\usepackage{cleveref} 
\glsdisablehyper
\usepackage[most]{tcolorbox} 

\usepackage{listings} 
\colorlet{punct}{red!60!black}
\definecolor{background}{HTML}{EEEEEE}
\definecolor{delim}{RGB}{20,105,176}
\colorlet{numb}{magenta!60!black}

\title{Beyond the Rabbit Hole:\\Mapping the Relational Harms of QAnon Radicalization}

\author{
  Bich Ngoc Doan \\
  EPFL \\
  Lausanne, Switzerland \\
  bich.doan@epfl.ch
  \And
  Giuseppe Russo \\
  EPFL \\
  Lausanne, Switzerland \\
  giuseppe.russo@epfl.ch
  \And
  Gianmarco De Francisci Morales \\
  CENTAI \\
  Turin, Italy \\
  gdfm@acm.org
}

\newenvironment{squishlist}
{\begin{list}{$\bullet$}
{\setlength{\itemsep}{0pt}
\setlength{\parsep}{3pt}
\setlength{\topsep}{3pt}
\setlength{\partopsep}{0pt}
\setlength{\leftmargin}{1.5em}
\setlength{\labelwidth}{1em}
\setlength{\labelsep}{0.5em}}}
{\end{list}}

\newcommand{\revise}[1]{#1}

\begin{document}
\maketitle
\begin{abstract}
Large-scale computational research on conspiracy theories has focused exclusively on believers' online behavior, leaving the harm experienced by those closest to them under-examined. This paper bridges this gap by analyzing \num{12747} stories from \texttt{r/QAnonCasualties}, an online support group for people who have ``lost'' someone to conspiracy beliefs. We design a computational pipeline to extract fine-grained thematic traits from personal narratives and cluster them into six coherent radicalization personas, which we then link to the emotional toll reported by narrators via LLM-assisted emotion detection and regression modeling. We find that personas are meaningful predictors of specific emotional harms: radicalization perceived as a deliberate ideological choice is associated with anger and disgust, while personas marked by personal and cognitive collapse correspond to fear and sadness. This work provides an empirically grounded computational framework for understanding the relational harms of radicalization, opening new avenues for research into its wider social consequences.
\end{abstract}

\section{Introduction}
Conspiracy theories pose a significant social and political challenge by driving polarization, eroding trust in public institutions~\citep{van2015conspiracy}, and inciting real-world violence~\citep{munn2021more}. As these theories proliferate online~\citep{russo2023spillover}, computational research has largely relied on social media analysis to map the diffusion of misinformation and track believers' online footprint~\citep{phadke2021social, wang2023identifying, engel2022characterizing, corso2025conspiracy}.

While this paradigm yields valuable large-scale insights into the online evolution of conspiratorial communities, it has two key limitations.
First, believer profiles are built primarily from online content, overlooking the offline characteristics that help explain the psychological vulnerabilities and triggers underlying belief adoption. Second, focusing on believers alone underestimates the consequences that reach beyond the individual to affect their immediate social circle. Anecdotal studies report the harm endured by relatives and acquaintances of people falling into the ``rabbit hole'': social erosion, depression, and the risk of adopting the same beliefs~\citep{xiao2021sensemaking, engel2023learning, mastroni2024one}. Yet these consequences remain unmeasured at scale, leaving a major blind spot in the literature.

To address this gap, we turn to \texttt{r/QAnonCasualties}, a subreddit where users post first-hand stories about relatives or close acquaintances who have adopted QAnon and adjacent conspiracy beliefs\revise{---arguably the largest and most prevalent contemporary case study of mass radicalization---}detailing the radicalization journey of the individual and the behavioral shifts observed throughout this process.
These data offer a dual perspective: each post captures the offline radicalization trajectory of the relative or acquaintance---whom we refer to as the \emph{object of narration} (OON)---as perceived by the post's author, defined hereafter as the \emph{narrator}, while simultaneously documenting the emotional and relational harm they experience.
To study this dual perspective, we formulate two research questions:

\begin{squishlist}
    \item[\textbf{RQ1:}] Which archetypes, or ``personas,'' emerge from the thematic traits perceived across the OON's radicalization trajectories?
    \item[\textbf{RQ2:}] How are these radicalization personas associated with the emotional toll experienced by the narrator?
\end{squishlist}

To answer \textbf{RQ1}, we use an LLM co-annotation framework to extract thematic traits from narrative texts and organize them—drawing on radicalization theory~\citep{borum2011radicalization}—into three temporal phases spanning pre-existing conditions, trigger factors and post-radicalization characteristics. We then model the co-occurrence of these traits to derive six data-driven radicalization personas. 

To address \textbf{RQ2}, we apply regression analysis to compositional data (via Isometric Log-Ratio transformation) to examine how exposure to these personas relates to narrators’ emotional outcomes. We link the OON's radicalization personas to reported psychological distress and demonstrate that they are strong predictors of specific emotional harm.

Overall, we make three contributions:

\begin{squishlist}
    \item[\textbf{C1:}] We release \texttt{QAnonCasualties-12k}, the first publicly available dataset 
    \revise{built from}
    \num{12747} first-hand stories documenting offline radicalization into QAnon, annotated with thematic traits characterizing the OON's radicalization trajectory and the narrator's emotional response.\footnote{We release Reddit post identifiers alongside our derived annotations (thematic traits, persona mixtures, emotion labels, and narrator control variables) rather than raw post text, to protect user privacy and respect Reddit's terms of service. Code and data are available at \url{https://github.com/ngoccc/qanoncasualties-analysis}.}
    \item[\textbf{C2:}] We derive six data-driven \emph{radicalization personas}, the first empirical typology of offline radicalization journeys as witnessed and reported by close relations.
    \item[\textbf{C3:}] We provide large-scale quantitative evidence that radicalization personas predict specific emotional harms experienced by the narrator, thus reframing radicalization as a measurable relational phenomenon.
\end{squishlist}

\section{Related Work}
\textbf{Characterizing Online Radicalization.}
Computational research has extensively mapped online radicalization through digital traces. Several works study how users are socially drawn into extremist communities~\cite{phadke2021social}, where echo chambers amplify divisive rhetoric~\cite{samory2018government}, violent content~\cite{rousis2022truth}, and emotionally-charged disinformation~\cite{sharma2022characterizing, marino2024polarization}. These communities also prove highly resilient, withstanding moderation by migrating to new platforms or spilling over into the mainstream~\cite{russo2023spillover, russo2023understanding}

Radicalization, however, extends beyond online spaces and reshapes the offline lives of its targets.
Susceptibility to conspiracy thinking~\citep{corso2025early} often builds on pre-existing vulnerabilities such as health struggles or institutional distrust~\cite{xiao2021sensemaking}, and may culminate in mental health challenges~\cite{engel2023learning}.
However, research documenting these dynamics relies mainly on small-scale interviews, which constrain generalizability.
Our study addresses this by analyzing the
narratives on \texttt{r/QAnonCasualties}, capturing the offline, observational dimensions of radicalization within a large corpus of first-person accounts.

\textbf{Radicalization Framework.}
Radicalization is a complex, multi-faceted process that varies across populations~\cite{wang2023identifying, phadke2022pathways}.
Several frameworks have been developed to model its components and progression.
At a granular level, research identifies radicalization through discrete signals: computationally, via the adoption of specific jargon or hateful language~\cite{fernandez2018understanding, rowe2016mining}, and qualitatively, via offline behaviors such as pathological fixation on public figures or explicit threats~\cite{reid2012role}.
These signals are typically part of a broader, incremental process.
Stage-based models further conceptualize radicalization as a temporal progression through phases, from grievance to indoctrination and action~\cite{borum2011radicalization, wolfowicz2021cognitive}.
We draw on these frameworks to operationalize radicalization as a trajectory of thematic traits, and examine how they coalesce into distinct personas (RQ1).

\textbf{Radicalization Impacts Beyond the Individual.}
Radicalization harms not only believers but also their social circle. 
There is growing evidence that having a loved one involved in a conspiracy theory may lead to anxiety and PTSD~\cite{moskalenko2023secondhand}, strain relationships~\cite{mastroni2024one}, and require specialized psychosocial support~\cite{st2023qollateral}.
The nature of these harms likely depends on their underlying causes, yet little is known about the spread of emotional tolls across radicalization profiles.
To fill this gap, our study links distinct radicalization personas to the form of distress \revise{with which they are associated}
(RQ2), thus providing the first large-scale test of this relation.

\section{QAnonCasualties-12k Dataset}

\subsection{Data Collection}
We construct our dataset from \num{31837} posts on \texttt{r/QAnonCasualties}, a subreddit where users document the radicalization of relatives or acquaintances into QAnon and adjacent conspiracies.
We focus on accounts that detail the believer's behavioral evolution over time.
Following \citet{antoniak2019narrative}, we preprocess the corpus by removing automated accounts, duplicates, and posts shorter than \num{50} words.
We then apply a structural filter to retain only those that capture an interpersonal dynamic: ones that mention both a \emph{narrator} (the post author) and an \emph{object of narration} (OON, the believer).
To operationalize this filter, we use a dependency parser~\citep{choi2015parser} to extract nouns and pronouns, manually categorize the most frequent items into a narrator set (e.g., ``I'', ``me'') and an OON set (e.g., ``mother'', ``husband''), as listed in \Cref{tab:entities}, and keep only posts containing at least one entity from each set.

\begin{table}[ht]
\footnotesize
\begin{center}
\caption{Entities and their representative n-grams.}
\label{tab:entities}
\begin{tabularx}{\columnwidth}{p{0.3\columnwidth} X}
\toprule
\textbf{Entity Category} & \textbf{n-grams} \\
\midrule
\multicolumn{2}{l}{\textit{Narrator}} \\[0.1em]
\hspace{0.5em} Narrator (self) & I, me, myself \\
\midrule
\multicolumn{2}{l}{\textit{Object of Narration (OON)}} \\[0.1em]
\hspace{0.5em} Parent & mother, father, parent \\
\hspace{0.5em} Sibling & brother, sister, sibling, cousin \\
\hspace{0.5em} Extended family & aunt, uncle, grandparent \\
\hspace{0.5em} Partner & husband, wife, boyfriend, girlfriend \\
\hspace{0.5em} Friend & friend, roommate, colleague \\
\bottomrule
\end{tabularx}
\end{center}
\end{table}

Three annotators reviewed a random sample of \num{200} retained posts: \num{87}\% satisfy the dual-perspective requirement (Fleiss’ $\kappa = 0.71$). The pipeline yields a final dataset of \num{12747} posts (additional details in Appendix~\ref{app:data}).

\subsection{Data Processing}
Posts on \texttt{r/QAnonCasualties} are unstructured: they mix biographical detail, affective testimony, and second-hand reporting, complicating the extraction of OON-specific attributes needed to reconstruct radicalization trajectories.
To address this, we apply BERTopic~\citep{grootendorst2022bertopic} to extract topical patterns from post sentences, using UMAP for dimensionality reduction and HDBSCAN for clustering (hyperparameters in Appendix~\ref{app:bertopic_config}).
This process yields \num{461} topics.
We benchmark BERTopic against LDA~\citep{blei2003latent} and prompt-based LLM topic modeling~\citep{wang2023prompting} (Appendix~\ref{app:bertopic_select}), and find that BERTopic achieves higher topic coherence, consistent with prior findings~\citep{egger2022topic, wang2023prompting}.

Manual inspection revealed that many of the \num{461} topics were not descriptive of OON attributes (e.g., trivial life details, conversational artifacts).
To filter them, we develop an \textbf{LLM Co-annotation Framework}: a human-in-the-loop pipeline combining LLM-assisted topic filtering with expert validation.
We curate few-shot exemplars of relevant and irrelevant topics—each with topic keywords, representative posts, a binary label, and a short rationale—and prompt \texttt{gpt-4o-mini} to classify each remaining topic, marking it relevant if it pertains to (i) the OON's belief adoption, (ii) behavioral change, or (iii) resulting interpersonal dynamics. The first author independently annotated all topics; disagreements with the LLM were resolved in consultation with the two co-authors, yielding a final set of \num{198} relevant topics, each subsequently labeled with a concise LLM-generated descriptor.

\section{RQ1: Trajectories and Personas}

\subsection{Radicalization Trajectories}
To characterize the OON's radicalization trajectories, we map the \num{198} OON-relevant topics onto the staged model of radicalization proposed by \citet{borum2011radicalization} and operationalized for empirical trajectory analysis by \citet{klausen2016behavioral}.
The model conceptualizes radicalization as a progression organized around a central catalyzing event (the \emph{trigger}) that mediates between an individual's antecedent vulnerabilities and their subsequent ideological and behavioral transformation.
This framing aligns naturally with our setting: testimonies on \texttt{r/QAnonCasualties} are typically structured along this temporal arc, recounting who their loved one was before (\emph{Pre-Radicalization Conditions}), what happened (\emph{Trigger}), and what they became after (\emph{Post-Radicalization Characteristics})---the three phases we adopt as the backbone of our analysis.

Each LLM-co-annotated topic is assigned to one of these phases by the first author.
Assignments were validated by the co-authors, with disagreements resolved through discussion.
We then manually merge conceptually similar topics 
into higher-level themes (Appendix~\ref{app:topics_agg}), yielding a final set of \num{50} thematic \emph{traits} of radicalization trajectories.
Each post is thus represented as a \emph{profile}: the subset of traits it expresses.

\Cref{fig:sankey-all} visualizes the resulting trajectory space as a flow across the three phases, with edge widths proportional to the frequency of co-occurring traits within profiles.
Three findings stand out:

\begin{figure*}[ht]
    \centering
    \includegraphics[width=\linewidth]{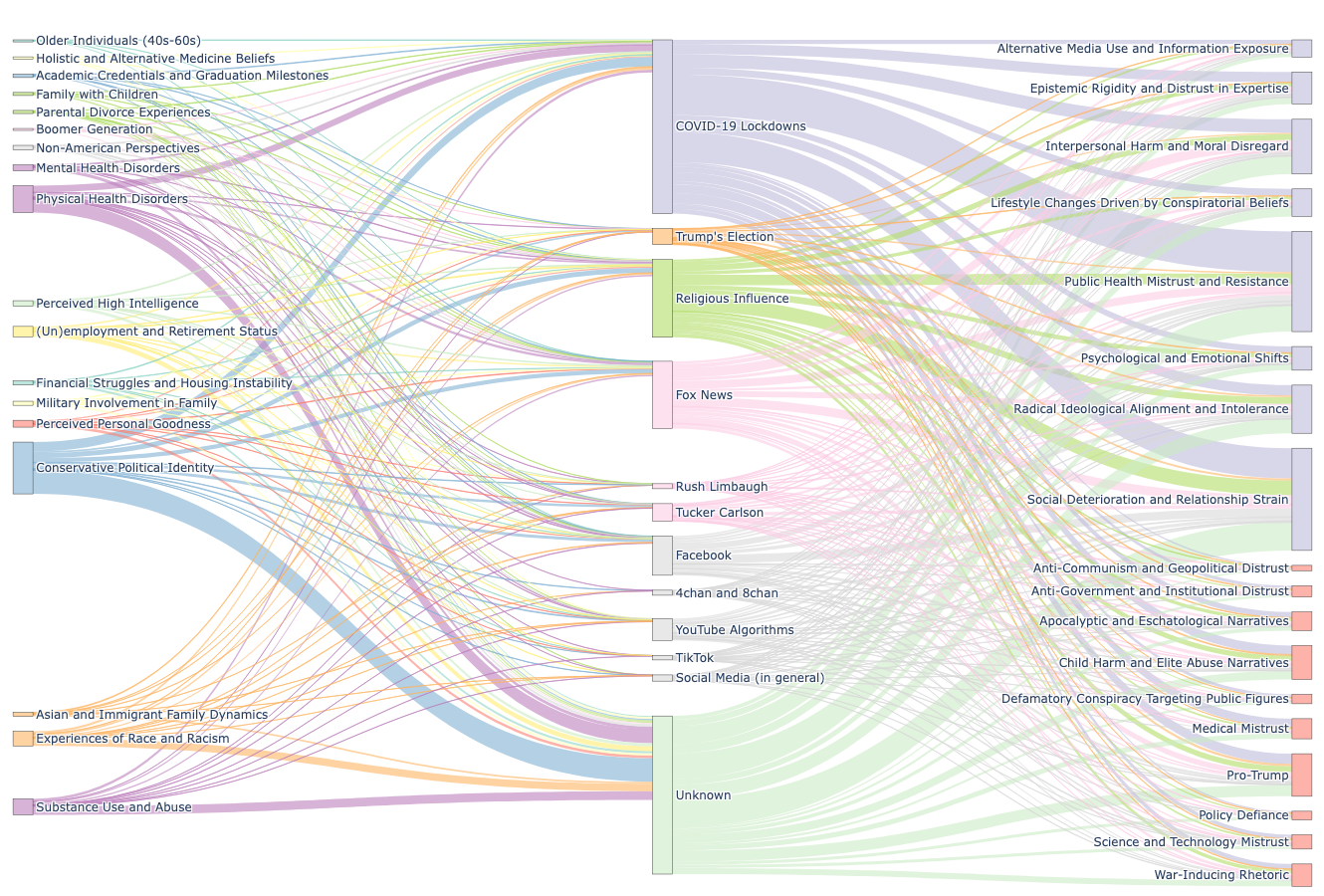}
    \caption{Flow of \num{50} thematic traits across the three radicalization phases.}
    \label{fig:sankey-all}
\end{figure*}

\paragraph{Radicalization antecedents are ideologically anchored but individually heterogeneous.}
\emph{Pre-Radicalization Conditions} span psychological (\verb|Physical Health Disorders|, \verb|Mental Health Disorders|), socio-economic (\verb|Financial Struggles|), and demographic dimensions.
Yet, \verb|Conservative Political Identity| stands out as the dominant antecedent, appearing in \num{1489} profiles---nearly $50$\% more than the next most common factor.
This suggests that, despite substantial heterogeneity in vulnerabilities, conservative ideological orientation is the strongest precursor to QAnon adoption in our corpus.

\paragraph{Triggers concentrate around a pandemic shock and its media ecosystem.} Trigger Factors are strikingly concentrated: \verb|COVID-19 Lockdowns| (\num{2402}) dominate by a wide margin, with the remainder clustering around \verb|Religious Influence| (\num{1181}) and platform-specific media exposure (\verb|Fox News|, \verb|Facebook|, \verb|YouTube Algorithms|).
Political events, by contrast, play a comparatively minor role (\verb|Trump's Election|, \num{181} profiles), pointing instead to the pandemic information environment as a potentially central driver of recruitment.
\revise{Nevertheless, a considerable share of profiles (1714) lack a clearly identifiable trigger (Unknown), suggesting the catalyzing moment is often internal or simply hidden from view.}

\paragraph{Post-Radicalization Characteristics cluster along two largely independent axes.}
Outcomes separate cleanly into a \emph{relational} axis (\verb|Social Deterioration| and \verb|Interpersonal Harm|) characterized by communication tension, isolation, and estrangement, together with shifts in \verb|Lifestyle| and \verb|Media Use|, and an \emph{ideological} axis (adoption of core QAnon tenets centered on \verb|Child Harm| and \verb|Apocalyptic Narratives|), interwoven with \verb|Pro-Trump| and \verb|War-Inducing Rhetoric|, and accompanied by a broader rejection of established knowledge via \verb|Medical| and \verb|Science and Technology Mistrust|.
The two axes co-occur in only \num{32.6}\% of profiles, thus suggesting that \emph{behavioral} and \emph{epistemic} radicalization are partially decoupled: a distinction that mirrors the two-pyramids account separating radicalization of action from radicalization of opinion~\citep{mccauley2017understanding}.


\subsection{Clustering Radicalization Profiles}
Building on the trajectory framework, we now ask whether the OON profiles cluster into a small number of recurring \emph{radicalization personas}, which we define as latent clusters capturing co-occurring sets of traits across profiles.

A key methodological assumption is that real-world radicalization profiles are unlikely to belong to a single archetype: a believer may, for instance, combine pandemic-era distrust of institutions with a pre-existing right-wing political identity and a religious worldview.
Hard-clustering methods such as $k$-means, which force each profile into a single cluster, would obscure precisely the kind of overlap we expect to find. We therefore adopt a probabilistic, mixed-membership formulation, in which each profile is represented as a distribution over personas and each persona as a distribution over traits.
Following \citet{yang2016did}, who use a similar formulation to derive behavioral archetypes, we instantiate this with Latent Dirichlet Allocation (LDA), representing each profile as a bag-of-traits.

To select the number of personas, we sweep $k$ from $4$ to $15$ and select $k$ via topic coherence (Appendix~\ref{app:persona_methods} and~\ref{app:persona_number}). The metric peaks at $k=6$, and a subsequent qualitative review of the resulting personas confirms that this value yields the most distinct and coherent archetypes. We use $k=6$ henceforth.

\subsection{Persona Analysis} \label{sec:persona_analysis}
Our six personas (\textbf{Health-Triggered Conspiracy Theorist}, \textbf{Political Extremist}, \textbf{Social Media Spiral}, \textbf{Religious Apocalypticist}, \textbf{Conservative Identity Protector}, and \textbf{Pandemic-Triggered Skeptic}) are described in \Cref{tab:personas_new}. Per-persona narratives and statistics are in Appendices~\ref{app:narr_comp} and~\ref{app:persona_stats}. Overall, these personas are not artifacts of hard partitioning: profiles occupy on average \num{1.70} personas (threshold $=1/k$); \num{52.2}\% exhibit mixed membership, confirming that the probabilistic formulation captures genuine overlap rather than serving as a methodological formality.


\input{tables/personas_new}

We frame the contribution of this typology in two parts.
First, \textbf{we provide the first integrated empirical typology of QAnon believers from witnessed testimony.}
Prior work has identified individual mechanisms in isolation: right-wing media ecosystems~\citep{sipka2022comparing}, algorithmic amplification~\citep{hannah2021conspiracy}, Christian eschatology~\citep{miotto2025stand}, COVID-19 conspiracy convergence~\citep{morelock2022nexus}, and personal health attribution to conspiratorial actors~\citep{xiao2021sensemaking}.
However, it has not shown how these mechanisms intertwine within a believer population.
Each of our six personas instantiates one of these previously documented mechanisms, and their relative frequencies (\Cref{tab:personas_new}) provide the first quantitative estimate of their respective prevalence in the QAnon-adjacent population observable through \texttt{r/QAnonCasualties}.
Second, \textbf{mixed membership reveals interaction effects invisible to single-mechanism studies.} The well-documented fusion of Christian symbolism with Trumpism~\citep{miotto2025stand} surfaces as co-loading on the Religious Apocalypticist and Political Extremist personas in over \num{1000} believers in our corpus. Mixed membership thus reveals not just who radicalized, but how distinct pathways converge and reinforce one another.


\section{RQ2: Persona-Specific Emotional Harm}
Our second goal is to quantify the impact that different radicalization personas have on the emotional toll experienced by their relatives and acquaintances, i.e., to model the relationship between the \emph{object of narration} (OON) and the \emph{narrator}.

\subsection{Regression Setup}
\label{sec:rq3_setup}
We model this relationship via a binary logistic regression.
For each of four negative emotions $y \in \{\text{anger, fear, sadness, disgust}\}$, we fit
\begin{equation}
\label{eq:logreg}
\mathrm{logit}\,\Pr(y=1 \mid \mathbf{x}, \mathbf{c}) = \beta_0 + \boldsymbol{\beta}^\top \mathbf{x} + \boldsymbol{\gamma}^\top \mathbf{c},
\end{equation}
where $y$ is the presence of the emotion in the narrative, $\mathbf{x}$ encodes the OON's persona composition, and $\mathbf{c}$ is a vector of narrator-level controls.
We describe each component in turn below.

\paragraph{Dependent variables: narrator emotions ($y$).}
We operationalize the emotional toll by using four discrete negative emotions from Plutchik's framework: anger, fear, sadness, and disgust~\citep{plutchik2001nature}.
We choose this categorical approach over dimensional models (e.g., valence and arousal) for its interpretability, as identifying specific emotions such as ``anger'' versus ``sadness'' provides more actionable insights into the nature of interpersonal harm than abstract affective scores.

Extracting these emotions from our corpus is non-trivial: posts are long-form, multi-actor, and describe highly specific lived experiences. Standard approaches, such as lexicon-based methods and pre-trained classifiers, fail to disentangle the narrator's emotions from those of the people they describe. Training a custom classifier would require costly manual annotation beyond the scope of this work. We therefore adopt an LLM-assisted strategy using \texttt{gpt-4o-mini}, selected for its strong instruction-following capabilities and cost efficiency. Prompts are iteratively designed and evaluated following prior LLM-based emotion detection work~\citep{greschner2024fearful, liu2024emollms, niu2024text}.
Empirical validation on a \num{100}-post sample confirms this choice: GPT-based annotation achieves mean $\kappa=0.6$ against two independent annotators, substantially outperforming both NRC~\citep{mohammad2013nrc} and GoEmotions-finetuned~\citep{lowe2022goemotions} baselines (full results \revise{and error analysis} in Appendix~\ref{app:emotion_ablation}). The final prompt (Appendix~\ref{app:prompt_emotion}) is applied to the full dataset, yielding four binary dependent variables.

\paragraph{Controls: narrator attributes ($\mathbf{c}$).}
To isolate the effect of persona composition, we control for three narrator-level attributes: the narrator's relationship to the OON, and their inferred gender and age.
We extract these attributes via a hybrid annotation pipeline standard in demographic inference~\citep{jagfeld2021understanding, rosario2023age, harvey2025using}.
First, we apply high-precision regular expressions adapted from \citealp{tigunova2020reddust} and our entity n-grams \Cref{tab:entities} (full details in Appendix~\ref{app:regex}) to identify explicit self-disclosures (e.g., ``my \emph{husband}'', ``I \emph{[35F]}'').
We then benchmark zero-shot classification with \texttt{gpt-4o-mini} against this set, after masking the explicit disclosures used as ground truth. The model achieves F1 score of $0.95$ for relationships, $0.77$ for age, and $0.70$ for gender.
Based on this validation, we use the LLM to infer the attributes for posts where they are not explicitly stated.

\paragraph{Predictors: persona composition ($\mathbf{x}$).}
The predictors of interest are the six radicalization personas identified in \Cref{sec:persona_analysis}.
For each post (i.e., profile), the LDA model estimates a probability distribution over the six personas, which would naturally serve as $\mathbf{x}$.
However, plugging these probabilities directly into \Cref{eq:logreg} is statistically invalid, and a transformation is required.

\paragraph{Handling Compositional Predictors via ILR.}
\label{sec:ilr_method}
The regression cannot be fit on the raw persona probabilities: they are non-negative and sum to one (\emph{compositional data}~\citep{aitchison1982statistical}), which violates the assumptions of standard linear regression.
We address this issue via an \emph{Isometric Log-Ratio} (ILR) transformation~\citep{egozcue2003isometric}, which maps the six probabilities into five orthogonal coordinates called \emph{balances}, each capturing the relative dominance of one group of personas (the \emph{numerator}) over another (the \emph{denominator}).
The grouping is defined by a Sequential Binary Partition (SBP) that reflects thematic distinctions across the personas: the top-level balance contrasts \emph{Situational} personas (radicalization stemming from external events such as a health crisis or algorithm) against \emph{Dispositional} ones (rooted in pre-existing identity or faith), with four further balances encoding nested contrasts within each group.
The full SBP and ILR formulation is shown in~\Cref{tab:sbp}. 

\input{tables/sbp}

\subsection{Regression Results}
\label{sec:rq3_results}
\begin{figure}[tb]
    \centering
    \includegraphics[width=\linewidth]{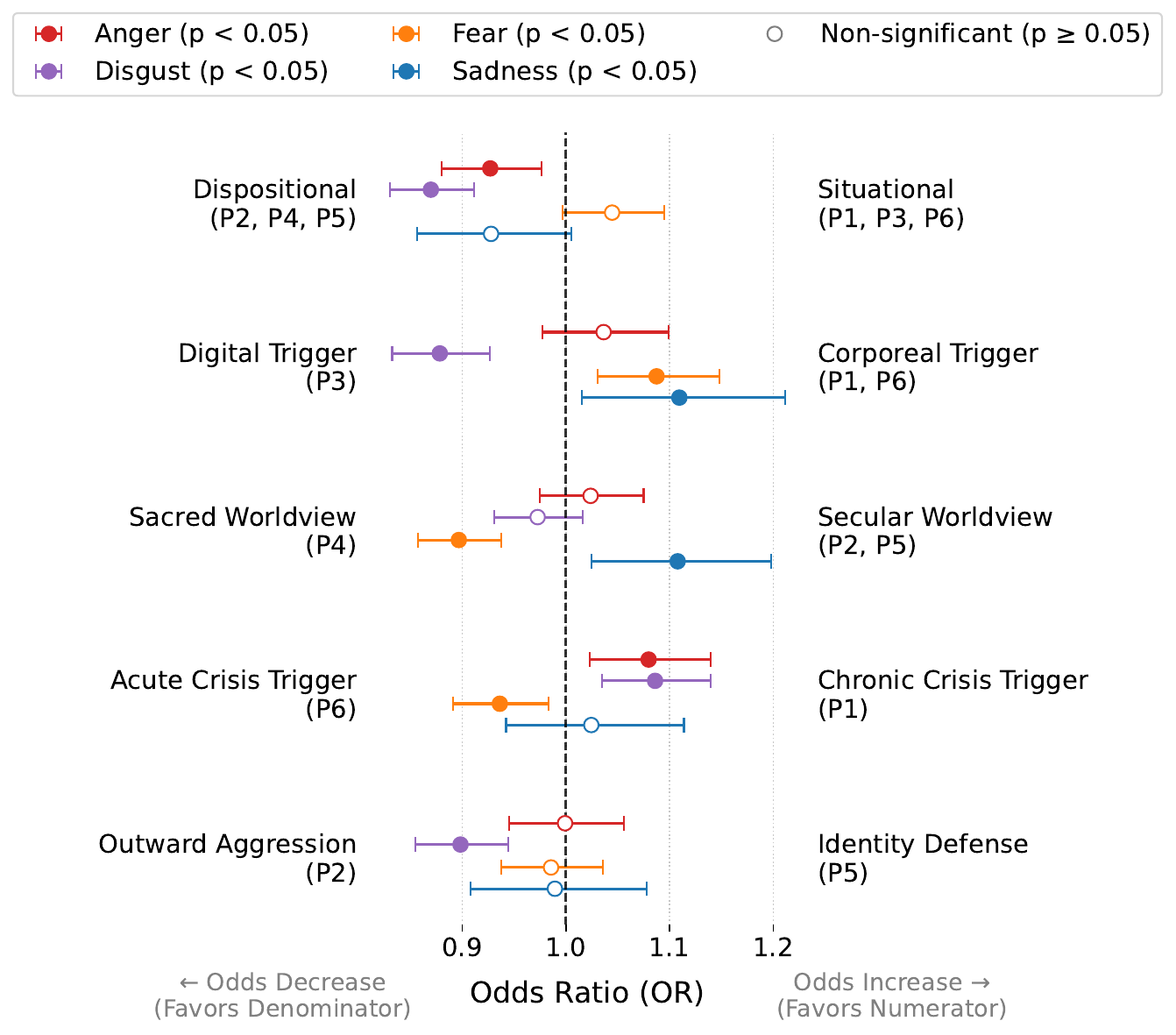}
    \caption{Effects of personas on narrator emotions. Points show odds ratios (OR) from logistic regression models, with bars indicating $95\%$ confidence intervals.}
    \label{fig:balances}
\end{figure}

We fit one logistic regression per emotion (\Cref{eq:logreg}) and report results as odds ratios (OR): $\text{OR}>1$ indicates higher odds of the emotion with greater dominance of the numerator group, whereas $\text{OR}<1$ indicates higher odds with the denominator group.
We interpret findings at two levels: (1) the high-level conceptual themes of each balance, and (2) the defining persona characteristics within those balances. The results, visualized in \Cref{fig:balances}, reveal two distinct emotional patterns.

\textbf{Anger and disgust are most strongly associated with radicalization perceived as an active process inflicting tangible harm rather than a passive misfortune}, mirroring psychological research linking anger to perceived intentionality~\citep{weiner1985attributional,vasilopoulos2019fear}.
This pattern emerges in two key balances.
First, our primary balance shows that profiles dominated by \emph{Dispositional} personas (rooted in pre-existing political or religious beliefs) are more likely to evoke anger (OR$=0.93$) and disgust (OR$=0.87$) than those driven by \emph{Situational} factors. 
Balance 4 refines this pattern within the Situational group: the \emph{Chronic Personal Crisis} persona (P1, characterized by destructive behaviors) is linked to more anger (OR$=1.08$) and disgust (OR$=1.09$) than the \emph{Acute Public Crisis} (P6, defined by rejection of scientific authority). Even when the initial trigger is external, directly harmful personal actions are seen as a more potent transgression than a more indirect, anti-authority stance.

Disgust, in particular, is sensitive to the specific moral content of the transgression~\citep{chapman2013things}. Balance 5 reveals that the \emph{Outward Aggression} persona (P2, defined by political extremism) corresponds to stronger disgust ($\text{OR}=0.90$) than the \emph{Inward Identity Defense} persona (P5, focused on identity-based anxieties).
Likewise, Balance 2 shows that the \emph{Digital-based} Social Media Spiral (P3), defined by an obsession with abhorrent child-harm narratives, exhibits higher odds of disgust ($\text{OR}=0.88$) than the \emph{Corporeal-based} personas (P1 and P6), which reflect more generalized personal and health-related deterioration.
These findings highlight that disgust may function as a targeted moral response to perceived ideological depravity, rather than as a generic reaction to harm.

\textbf{Fear and sadness are linked to grieving a perceived cognitive and personal collapse}—a form of ``ambiguous loss''~\citep{pauline2009ambiguous} where narrators mourn a loved one's profound psychological transformation and loss of identity.
The tangible evidence of a loved one's physical and mental decline in the \emph{Corporeal-based} personas (P1, P6) has a stronger association with both fear ($\text{OR}=1.09$) and sadness ($\text{OR}=1.11$) than the more abstract \emph{Digital-based} projection of P3 (Balance 2).
Moreover, the ideological framing of the radicalization reveals a divergent impact on these emotions.
Balance 3 shows fear to be more prominent in narratives of the \emph{Sacred} Religious Apocalypticist (P4) ($\text{OR}=0.90$), while sadness is more prevalent in those of \emph{Secular} personas (P2, P5) ($\text{OR}=1.11$).
This nuanced result suggests that while religious detachment from reality is profoundly frightening, narrators perceive the hostility-induced transformation of secular extremism as a more saddening loss of identity---echoing the interpersonal grief documented in studies of affective polarization~\citep{iyengar2019origins}.

\section{Discussion and Conclusion}
This study introduced and validated a computational framework for moving beyond the individual conspiracy believer to model the offline, relational consequences of online radicalization.
By analyzing thousands of personal stories, we first developed a pipeline to map radicalization trajectories and derive six recurring radicalization personas.
We then provided the first large-scale, quantitative evidence that these computationally-derived personas are strong predictors of the specific emotional harms inflicted on loved ones.

A core strength of our multi-stage pipeline---combining contextual topic modeling, graphical modeling, and LLM-assisted classification---is its ability to extract multiple components of narrative structure from unstructured text~\cite{piper2021narrative}.
Its modular design makes it adaptable to other narrative-driven phenomena—such as other belief-based communities or support group dynamics, where similar pipeline stages could be selectively applied to extract narrative components of interest. \revise{Such adaptation would require re-deriving the domain-specific components such as entity n-grams, thematic traits, and LLM prompts for the new setting, while the pipeline's architecture, from temporal mapping through persona aggregation to outcome regression, remains directly reusable.}

Beyond methodology, the six personas represent a foundational feature set: by capturing stable, archetypal patterns in how radicalization unfolds, they open avenues for predictive modeling.
For example, future work could explore the use of these personas to forecast critical outcomes such as relationship severance, the likelihood of offline violence, or one's receptiveness to de-radicalization.
In doing so, it takes a step toward measuring not just who radicalizes, but also what radicalization leaves behind.

\clearpage

\section*{Limitations and Future Work}
Our study has several limitations that highlight avenues for future work.
First, our operationalization of emotional toll, while providing a clear and interpretable starting point, relies on pre-defined negative emotions.
This simplification inevitably overlooks more complex, context-specific feelings prevalent, such as helplessness, irony, or betrayal.
This shortcoming lays the groundwork for future research to develop a more nuanced data-driven taxonomy of the emotional harms associated with radicalization. 
\revise{In addition, our reliance on LLMs paired with limited human validation represents a trade-off between scale and precision, introducing errors from both sides: from the LLM itself, and from the narrow scope of human review, which was largely disagreement-triggered rather than exhaustive. A promising avenue is to scale up human validation, which would also provide a stronger empirical basis for developing classifiers tailored to radicalization-related narratives, better able to capture subtle linguistic markers of trauma and support-seeking than general-purpose models.}
Finally, the generalizability of our findings is bounded by 
our dataset's specificity. Drawing exclusively from \texttt{r/QAnonCasualties} introduces self-selection bias, where narrators seeking online peer support may not represent all those affected by radicalization,
\revise{influencing which pathways appear more or less common in our analysis.}
Future work should address this along three directions: extending to other platforms and media to mitigate selection effects, analyzing data longitudinally to model how radicalization trajectories evolve over time, and testing whether our personas generalize across other conspiracy movements beyond QAnon.

\section*{Ethical Consideration}
The data used in this study were sourced from the public subreddit \texttt{r/QAnonCasualties}. Our collection and analysis of this data for non-commercial research purposes are consistent with Reddit’s Terms of Service. While the content is publicly accessible, we take additional steps to protect the privacy of both the narrators (the post authors) and the objects of narration (the radicalized individuals they describe). All data was anonymized, and we removed any personally identifiable information (PII) such as usernames. All quantitative results presented in this paper are in an aggregated form. We report on collective patterns (e.g., persona distributions, odds ratios). The anonymized text fragments sent to the LLM APIs for annotation contained no personal information. We complied with the terms of service of the LLM provider for all analyses.

\bibliography{custom}

\renewcommand\thefigure{\thesection.\arabic{figure}}
\renewcommand\thetable{\thesection.\arabic{table}}
\setcounter{figure}{0}
\setcounter{table}{0}
\input{appendix}
\end{document}

%% file: tables/personas_new.tex
\begin{table*}[!ht]
\renewcommand{\arraystretch}{0.7}
\scriptsize
\begin{center}
\caption{The six radicalization personas, their core descriptions, and their dominant trajectory sub-diagrams.}
\label{tab:personas_new}
\begin{tabular}{>{\centering\arraybackslash}m{0.2cm} >{\raggedright}m{1.3cm} m{6cm} m{6.8cm}}

\toprule
\textbf{No.} & \textbf{Persona} & \textbf{Description} & \textbf{Sub-Sankey Diagram} \\

\midrule
P1 & The Health-Triggered Conspiracy Theorist & This persona's radicalization begins with a personal health crisis or substance abuse issue. Their initial mistrust of doctors evolves into a comprehensive conspiratorial worldview that eventually incorporates QAnon. This pathway is marked by severe social and professional collapse, including psychotic episodes, aggressive behavior, and the destruction of their former relationships and career. &  \adjustbox{valign=c}{\includegraphics[width=\linewidth]{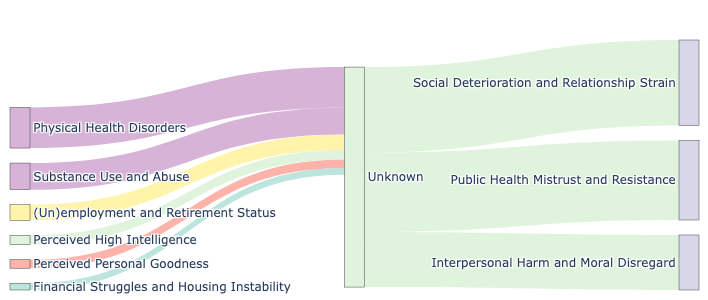}} \\ \midrule
P2 & The Political Extremist        & 
Driven by a pro-Trump identity and a diet of right-wing media, this persona adopts a fierce ``us versus them'' mentality. They view political opponents as an existential threat, justifying extreme rhetoric and the destruction of personal relationships over political disagreements. While they may carefully vet sources within their media bubble, they remain trapped in an echo chamber that reinforces their apocalyptic fears. &  \adjustbox{valign=c}{\includegraphics[width=\linewidth]{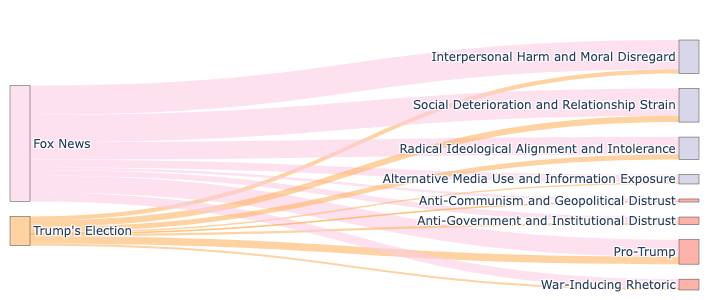}} \\ \midrule
P3 & The Social Media Spiral      & This persona falls down algorithmic rabbit holes on platforms such as YouTube and 4chan. What starts as a simple search for alternative information becomes a spiral of increasingly extreme content, culminating in an obsession with child harm narratives. They lose the ability to think critically and become highly susceptible to the influence of online influencers, while rejecting mainstream sources. & \adjustbox{valign=c}{\includegraphics[width=\linewidth]{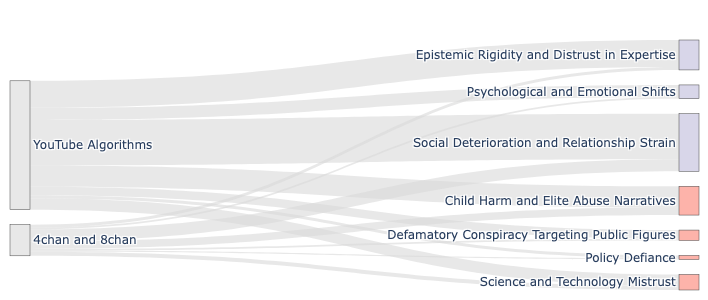}} \\ \midrule
P4 & The Religious Apocalypticist       & This persona fuses conservative Christianity with political conspiracy, interpreting current events through the lens of Biblical end-times prophecy. QAnon is seen as proof of a divine spiritual war, creating a powerful, self-reinforcing belief system. This worldview often includes a fear of technology as the ``mark of the beast'' and a conviction that a final battle between good and evil is imminent. & \adjustbox{valign=c}{\includegraphics[width=\linewidth]{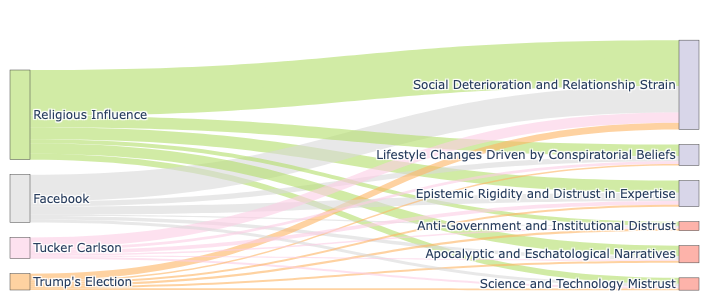}} \\ \midrule
P5 & The Conservative Identity Protector & This persona becomes radicalized in what they see as a defense of traditional conservative values against cultural threats, such as changing norms around race, gender, and sexuality (e.g., LGBTQ+ movements). Their identity becomes so paramount that they view disagreement as a personal betrayal, leading to an intolerance for opposing views and a willingness to prioritize ideological purity over family relationships. & \adjustbox{valign=c}{\includegraphics[width=\linewidth]{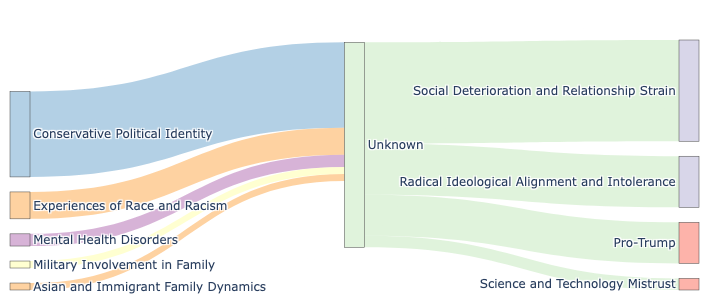}} \\ \midrule
P6 & The Pandemic-Triggered Skeptic             & Ignited by the COVID-19 pandemic, this persona’s pre-existing skepticism of mainstream medicine and preference for holistic health made them vulnerable to anti-vaccine messaging. This initial doubt quickly expands into broader government control conspiracies and a complete rejection of medical and scientific authority. &  \adjustbox{valign=c}{\includegraphics[width=\linewidth]{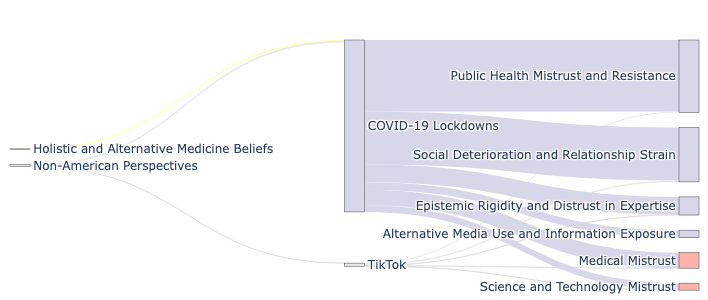}}                       
\\
\bottomrule
\end{tabular}
\end{center}
\end{table*}

%% file: tables/sbp.tex
\begin{table*}[t]
\footnotesize
\centering
\caption{Sequential Binary Partition (SBP) and Interpretation of the ILR Balances.}
\label{tab:sbp}
\begin{tabularx}{\textwidth}{
    >{\raggedright\arraybackslash}p{1.3cm}
    >{\raggedright\arraybackslash}p{8.3cm}
    >{\raggedright\arraybackslash}p{2.2cm}
    >{\raggedright\arraybackslash}p{2.2cm}
}
\toprule
\textbf{Balance \#} & \textbf{Conceptual Grouping and Rationale} & \textbf{Numerator Group} & \textbf{Denominator Group} \\
\midrule
1 & \textit{Situational vs. Dispositional:} This balance separates personas where radicalization is attributed to external, situational factors (e.g., a health crisis, algorithm) from those attributed to one’s internal, pre-existing disposition (e.g., their political identity or religious faith). & P1 (Health), P3 (Social Media), P6 (Pandemic) & P2 (Political), P4 (Religious), P5 (Conservative) \\
\midrule
2 & \textit{Corporeal vs. Digital Trigger:} Within the Situational group, this balance distinguishes between entry points rooted in the physical body and health (Corporeal) versus those rooted in the abstract, algorithmic world of online platforms (Digital). & P1 (Health), P6 (Pandemic) & P3 (Social Media) \\
\midrule
3 & \textit{Secular vs. Sacred Worldview:} Within the Dispositional group, this balance separates radicalization based on modern political identities (Secular) from one based on eschatological, faith-based narratives (Sacred). & P2 (Political), P5 (Conservative) & P4 (Religious) \\
\midrule
4 & \textit{Chronic vs. Acute Crisis:} Within the Corporeal group, this balance separates radicalization stemming from a long-term, chronic personal crisis (P1) from one triggered by a specific, acute public crisis (P6). & P1 (Health) & P6 (Pandemic) \\
\midrule
5 & \textit{Inward Identity Defense vs. Outward Aggression:} Within the Secular group, this balance separates an inward-looking identity defense (protecting conservative values) from explicit outward aggression (``us vs. them'' political combat). & P5 (Conservative) & P2 (Political) \\
\bottomrule
\end{tabularx}
\end{table*}

%% file: appendix.tex
\appendix
\section{Appendix}
\subsection{Data Preparation Details} \label{app:data}
\begin{figure}[ht!]
    \centering
    \includegraphics[width=\linewidth]{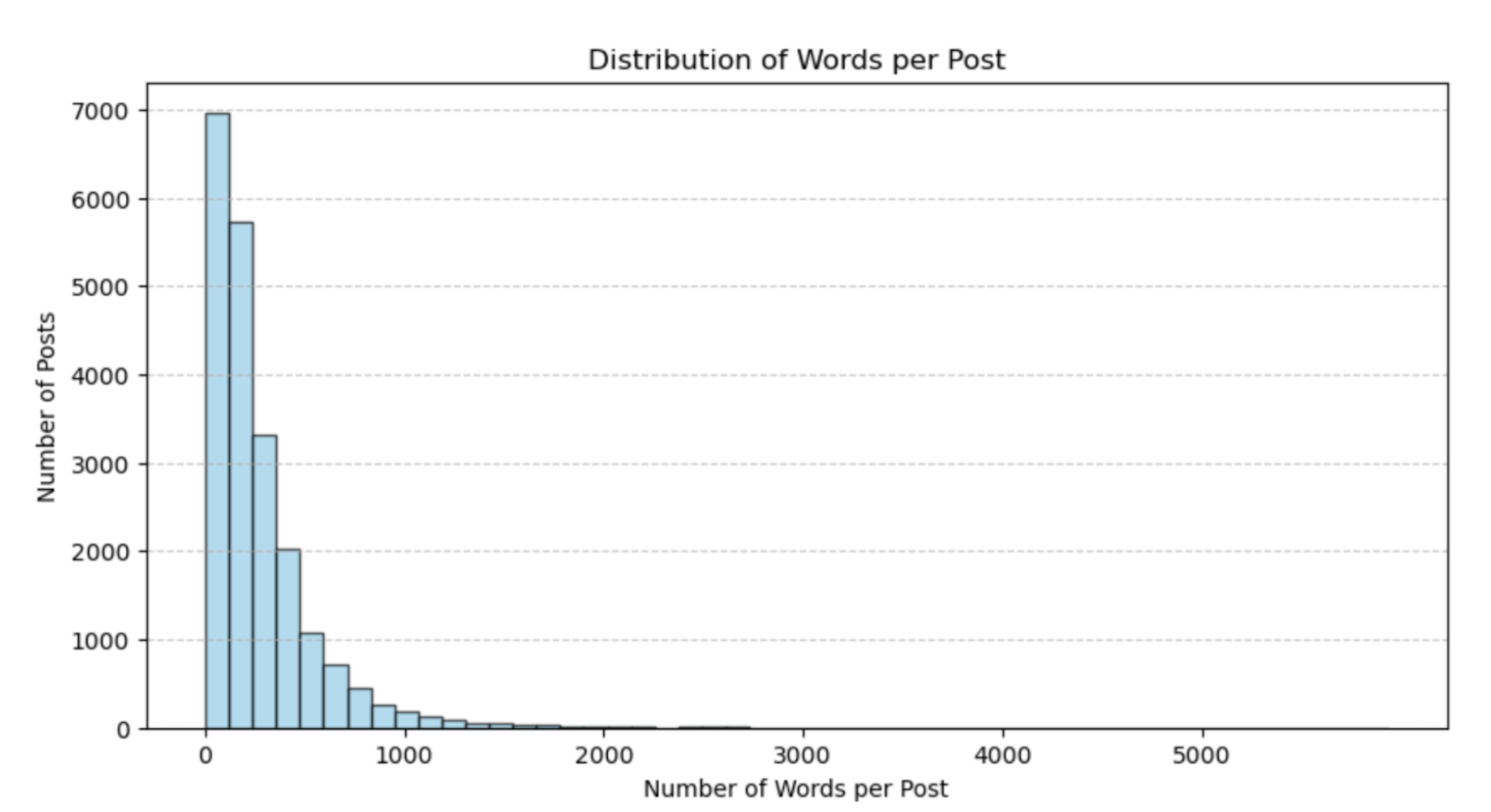}
    \caption{Distribution of post lengths (word counts).}
    \label{fig:post_length}
\end{figure}

Here we elaborate on the data preparation steps summarized in the main text. The data obtained covers the period between its launch on July 3, 2019, and December 31, 2024.
In the primary filtering stage, we exclude moderator and bot content by removing posts with the \texttt{distinguished} field set to \emph{moderator}, authored by the \texttt{AutoModerator}, or containing polls (via the \texttt{poll\_data} field).
We also exclude posts self-labeled with non-narrative flairs such as `Off-topic' or `Media Request'. Duplicate posts are identified and removed by exact match on post text (\texttt{selftext} field).
We choose the 50-word length threshold for removing short posts after examining the overall post length distribution (see~\Cref{fig:post_length}), where we manually audit samples at different lengths to ensure the cutoff removes trivial content while retaining concise but complete narratives.

The second stage extracts entities to enforce the dual-perspective structure described in the main text. While \Cref{tab:entities} reports representative n-grams for conciseness, \Cref{tab:entities_full} provides the complete set of n-grams used for filtering, including spelling variants and synonyms (e.g., ``mom'', ``mum'', and ``mother'' are all mapped to the Parent category).

\begin{table}[h]
\begin{center}
\scriptsize
\caption{Entities and their full n-grams.}
\label{tab:entities_full}
\begin{tabular*}{\columnwidth}{@{\extracolsep{\fill}} l p{0.7\columnwidth}}
\toprule
\textbf{Entity} & \textbf{n-grams} \\
\midrule
Narrator (self) & I, me, myself \\
Parent & mom, mum, mother, dad, father, parent \\
Sibling & brother, sister,  sibling, cousin \\
Extended family & grandma, grandpa, grandfather, grandmother, grandparent, aunt, uncle \\
Partner & husband, wife, boyfriend, girlfriend, spouse, fiance, partner \\
Friends & friend, roommate, colleague \\
\bottomrule
\end{tabular*}
\end{center}
\end{table}


\subsection{BERTopic Model Selection} \label{app:bertopic_select}
We conduct an empirical comparison of three topic modeling approaches for our dataset: BERTopic, LDA, and an LLM-based method following the PromptTopic pipeline~\cite{wang2023prompting} with \texttt{gpt-4o-mini}. All models are evaluated on the full filtered set of \num{12747} posts with $k=50$ topics fixed across models. We assess topic quality via NPMI coherence and diversity, and stability via Adjusted Rand Index (ARI) across three independent runs.

As shown in \Cref{tab:model_selection}, BERTopic outperforms both alternatives across all metrics. Most notably, its stability (ARI = 0.87) substantially exceeds that of LDA (0.14) and the LLM-based approach (0.53), indicating that BERTopic produces considerably more consistent topic assignments—a critical property for downstream persona derivation. We therefore proceed with BERTopic and tune its hyperparameters as described below.

\begin{table}[h]
\centering
\small
\caption{Topic quality and stability across modeling approaches ($k=50$, three independent runs).}
\label{tab:model_selection}
\begin{tabular}{lccc}
\toprule
\textbf{Method} & \textbf{Diversity} & \textbf{Coherence} & \textbf{ARI} \\
\midrule
BERTopic & \textbf{0.700} & \textbf{-0.030} & \textbf{0.873} \\
LDA & 0.317 & -0.064 & 0.135 \\
LLM-based & 0.092 & -0.136 & 0.533 \\
\bottomrule
\end{tabular}
\end{table}

\subsection{BERTopic Configuration} \label{app:bertopic_config}
We begin with BERTopic's default UMAP and HDBSCAN parameters and tune iteratively, evaluating topic quality via hierarchical clustering dendrograms and intertopic distance maps. We observe that default settings produce coherent but overlapping topics-reflected in densely clustered regions of the intertopic distance map-and incrementally increase \texttt{min\_cluster\_size} until overlap is substantially reduced. The research team assessed coherence at each iteration, with disagreements resolved by consensus. The final configuration sets HDBSCAN's \texttt{min\_cluster\_size} to 50, with all other parameters at their defaults (see \Cref{tab:bertopic-params}).
HDBSCAN identified \num{128667} outlier sentences (\num{35.3}\%), which were discarded, retaining \num{235694} sentences for subsequent analysis.
\begin{table}[h]
\centering
\small
\caption{BERTopic configuration parameters.}
\label{tab:bertopic-params}
\begin{tabular}{ll}
\toprule
\textbf{Parameter} & \textbf{Value} \\
\midrule
UMAP \texttt{n\_neighbors} & 15 \\
UMAP \texttt{n\_components} & 5 \\
UMAP \texttt{metric} & cosine \\
HDBSCAN \texttt{min\_cluster\_size} & 50 \\
HDBSCAN \texttt{metric} & euclidean \\
\bottomrule
\end{tabular}
\end{table}

\subsection{Merging Radicalization Topics} \label{app:topics_agg}
In the first research question, where we examine radicalization topics, we perform an aggregation process that consolidates 198 initial BERTopic outputs into 50 final themes used for the trajectory analysis.
The rationale behind this choice is that when assigning the 198 topics to our three-phase framework, we observe a strong concentration in Phase 3 (\emph{Post-Radicalization Characteristics}).
This distribution likely reflects a narrative focus on the tangible, observable consequences of radicalization, which are often more salient to the storyteller than the specific pre-existing conditions or triggers.
While this data distribution is itself an interesting finding, the high density of fine-grained topics in Phase 3 poses a challenge for creating a clear and interpretable visualization of the radicalization trajectories.
A Sankey diagram with a disproportionately large and granular final stage would obscure the broader patterns of flow.

To address this and improve analytical clarity, we perform the thematic aggregation step, focusing on the Phase 3 topics.
The full hierarchical structure of this consolidation for Phase 3 is presented in \Cref{tab:phase3_topics}.
\revise{We additionally introduced "Unknown" as a label during visualization to represent profiles whose extracted topics contain no Phase 2 (Trigger) content, i.e. only Phase 1 and Phase 3 topics were identified. The main paper's Sankey diagram (\Cref{fig:sankey-all}) displays this along with the final consolidated themes.}

\input{tables/phase3-topics}

\subsection{LLM Prompts}
This section details all the prompts we used with the \texttt{gpt-4o-mini} for LLM annotation.
\lstset{
  basicstyle=\ttfamily\small,
  breaklines=true,
  frame=single,
  columns=fullflexible,
  keepspaces=true,
  numbers=none
}
\subsubsection{Radicalization Topic Relevance Classification}
\label{app:prompt_rel}
\begin{lstlisting}[language={}]
You are a qualitative coding assistant helping analyze radicalization pathways from r/QAnonCasualties posts.

Determine whether the following topic is relevant to understanding the Object of Narration's (OON's) radicalization pathway.

Give Label: Yes/No and Reason.

{few_shots}
---
Topic representation: {representation}
Representative docs: {representative_docs}
Label:
\end{lstlisting}

\subsubsection{Radicalization Topic Labeling}
\label{app:prompt_label}
\begin{lstlisting}[language={}]
You are analyzing a topic derived from the r/QAnonCasualties subreddit, where individuals share personal stories about their relatives' involvement in the QAnon conspiracy. These narratives often include emotional experiences, opinions, and real-world context.
Topic Keywords: {representation}
Representative Documents: {representative_docs}

Please provide a concise and descriptive topic label depicting the main theme of the above content. Be specific and don't infer any additional information that is not stated in the text. Be action-oriented and focus on the object of narration. Be concise, straightforward and no need to mention QAnon context in the label.

If the topic is unclear or contains multiple equally prominent themes, respond with:
Unsure - <Reason>

Format: <Concise Label>
\end{lstlisting}

\subsubsection{Control Variable Extraction}
\label{app:prompt_control}

For relationship:
\begin{lstlisting}[language={}]
You are a helpful assistant analyzing Reddit posts from the subreddit r/QAnonCasualties. Your task is to identify the relationship of the person they are describing (the object of narration, or OON, who radicalized) to the narrator (the person writing the post), if such a relationship exists.

If the narrator is telling a personal story about someone they know who became radicalized by QAnon, output the precise relationship label (e.g., "mother", "father", "brother", "sister", "wife", "friend").
If the post is not about a specific person known to the narrator (e.g., general discussion, support, resource sharing), or if you are not sure, return: "N/A".
If the post talks about a stranger or a person not clearly connected to the narrator, return: "stranger".
If the post talks about the radicalization/recovery story of the narrator themselves, return: "self".
If multiple relationships are discussed, return the closest and most central one, prioritizing immediate family or partner.

Input:
Title: {title}
Body: {selftext}

Output:
Relationship:
\end{lstlisting}

For age group:
\begin{lstlisting}[language={}]
You are a helpful assistant analyzing Reddit posts from the subreddit r/QAnonCasualties. Your task is to estimate the narrator's age group, which is either "adolescence" (13-20) or "adult" (21+).
If unsure, return: "N/A".

Input: "{text}"

Output:
\end{lstlisting}

For gender:
\begin{lstlisting}[language={}]
You are a helpful assistant analyzing Reddit posts from the subreddit r/QAnonCasualties. Your task is to estimate the narrator's gender, which is either "f" (female) or "m" (male).
If unsure, return: "N/A".

Input: "{text}"

Output:
\end{lstlisting}

\subsubsection{Emotion Classification}
\label{app:prompt_emotion}
\begin{lstlisting}[language={}]
You are an emotionally-intelligent and empathetic agent.
You will be given a piece of text derived from posts in the subreddit r/QAnonCasualties, which are personal narratives describing one's experiences with a loved one who has been radicalized.
You must identify the emotions explicitly stated or clearly implied by the writer about their own feelings.
Only consider the writer's emotions - ignore the emotions of other people mentioned.
If the emotion is expressed at least once in the text, output 1; otherwise, output 0.
Do not infer emotions based solely on the topic or context unless the writer states or clearly implies them.
Possible emotions: Anger, Disgust, Sadness, Fear, Surprise.

Text: "{text}"

Please provide outputs for these emotions in JSON format:
{{
    "anger": 0 or 1,
    "fear": 0 or 1,
    "sadness": 0 or 1,
    "disgust": 0 or 1,
    "surprise": 0 or 1,
}}`
    
\end{lstlisting}

\subsection{Exploring Methods for Identifying Personas}
\label{app:persona_methods}
Our trajectory analysis, visualized in the main paper's Sankey diagram (\Cref{fig:sankey-all}), reveals a complex system with a high degree of interconnectedness between thematic traits.
We observe no clear, hard boundaries that would suggest mutually exclusive clusters of radicalization journeys.
This data structure strongly indicates that hard clustering methods such as k-means are ill-suited, as they would force each individual's profile into a single, exclusive archetype, failing to capture the blended nature of their experiences.
This observation also aligns with the theoretical understanding of radicalization as a complex and multifaceted phenomenon where different ideological and personal pathways often converge.
Therefore, we seek a soft clustering method that allows for mixed membership.
This choice narrows our evaluation to two primary candidates: NMF, which provides an additive decomposition, and LDA, which offers a probabilistic mixture.

\begin{figure}[h!]
    \centering
    \includegraphics[width=0.9\linewidth]{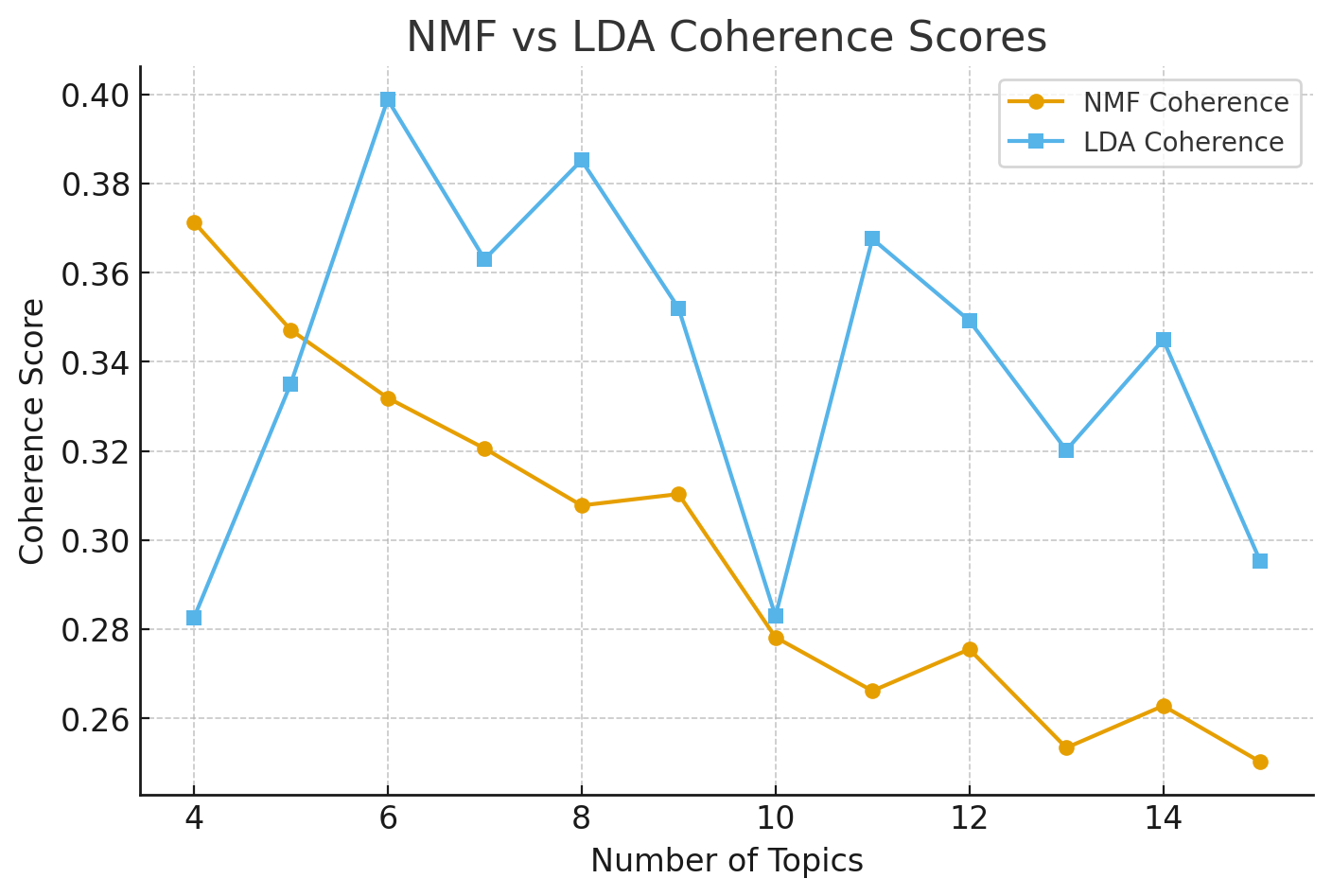}
    \caption{NMF vs LDA coherence score.}
    \label{fig:cvs}
\end{figure}

To empirically compare LDA and NMF, we train models for both methods with the number of personas (\texttt{k}) ranging from 4 to 15. 
We compare the models using the \texttt{C\_v} topic coherence score.
As shown in \Cref{fig:cvs}, the LDA models generally achieves higher coherence scores than the NMF ones.
Manual inspection on a selected few samples also verifies the superior interpretability of the LDA approach. 
Based on this convergent evidence, we select the LDA-inspired model for our final analysis.

\subsection{Selection of Persona Number} \label{app:persona_number}

As seen in the coherence score plot (\Cref{fig:cvs}), the score for LDA peaks at $k=6$.
While this provides a strong quantitative signal, the coherence scores for $k=8$ and $k=11$ are also high, necessitating a qualitative review to select the most meaningful and non-redundant set of personas.

\begin{figure}[ht!]
    \centering
    \begin{subfigure}{\linewidth}
        \centering
        \includegraphics[width=0.8\linewidth]{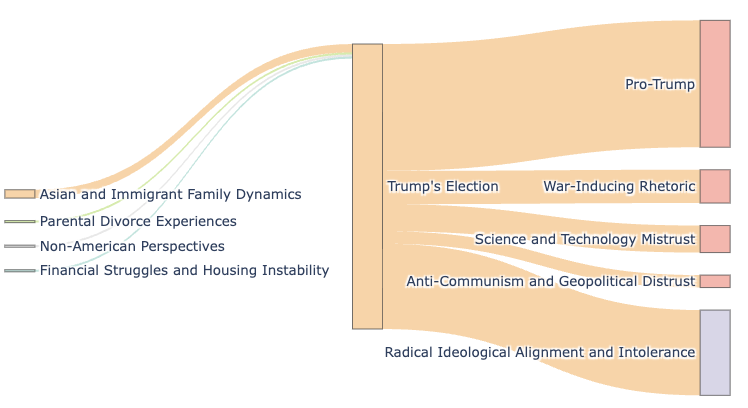}
        \label{fig:example1}
    \end{subfigure}
    \begin{subfigure}{\linewidth}
        \centering
        \includegraphics[width=0.8\linewidth]{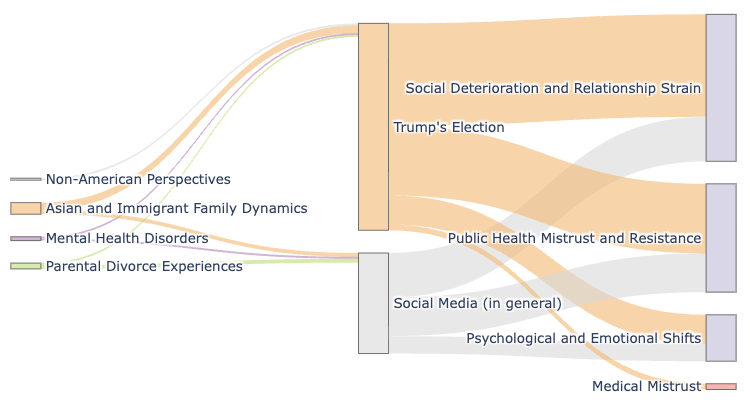}
        \label{fig:example2}
    \end{subfigure}
    \caption{An example of a ``fractured'' persona from the $k=8$ LDA model.}
    \label{fig:fractured_personas}
\end{figure}

We then conduct a qualitative review of the models with the highest coherence scores ($k=6, 8$, and $11$), examining the top traits, sub-Sankey diagrams, and representative narratives for each persona.
Our review reveals that models with higher k values often produces ``fractured'' personas, in which a coherent archetype is unnecessarily split into multiple, redundant sub-personas.
For example, as illustrated in \Cref{fig:fractured_personas}, the $k=8$ model generates two personas centered on the Trump Election trigger.
One emphasizes radical ideologies and anti-authority themes, and the other medical distrust, yet both are minor variations of the same overarching Political Extremist archetype already present in the $k=6$ model.
This redundancy would dilute clarity without adding new insight.
In contrast, the $k=6$ model produces the most distinct and coherent personas, with each representing a unique radicalization pathway and minimal thematic overlap.
Based on this convergent evidence---the peak coherence score at $k=6$ and its superior qualitative distinctiveness and non-redundancy---we select $k=6$ for the final analysis.

\subsubsection{Persona Stability Analysis} \label{app:persona_stability}
We conducted a stability analysis of the final $k=6$ persona model by re-fitting LDA five times with different random seeds ($42, 123, 456, 789, 999$). Since persona ordering is arbitrary across independent LDA runs, we first aligned personas across runs using the Hungarian algorithm on topic-profile ($\beta$) similarity, then evaluated stability of both the persona-trait distributions ($\beta$) and document-persona allocations ($\theta$) using this fixed alignment. We note that cosine similarity on $50$-dimensional trait distributions is inherently sensitive to run-to-run variation in the low-probability tail of the distribution.

Across the five runs, matched personas showed an average cosine similarity of $0.639$ (SD = $0.219$) for topic-profile ($\beta$) distributions and $0.667$ (SD = $0.139$) for document-topic ($\theta$) allocations. The coherence score that determines $k=6$ varied minimally across seeds (SE = 0.015).

Stability was not uniform across the six personas, as shown in Table~\ref{tab:persona_stability}. Persona 6 (Pandemic-Triggered Skeptic) was the most stable by a wide margin (cosine similarity = $0.956$, SD = $0.034$), while Persona 3 (Social Media Spiral) showed the lowest average similarity ($0.409$)---though notably, its low standard deviation ($0.067$) indicates this is a \emph{consistently} lower-cosine match across seeds rather than an erratic one. The remaining four personas fall in a moderate mid-range ($0.591$--$0.712$); of these, Persona 1 (Health-Triggered Conspiracy Theorist) has the second-highest variability (SD = $0.167$), suggesting it is not fully immune to run-to-run instability despite its higher mean similarity. We report this per-persona variation transparently and note it as a limitation: results involving lower-stability personas, particularly Persona 3, should be interpreted with correspondingly more caution.

\begin{table}[t]
\centering
\caption{Per-persona stability: average cosine similarity (topic-profile $\beta$) between matched personas across five LDA re-fits with different seeds. Persona numbers correspond to Table~\ref{tab:personas_new}.}
\small
\begin{tabular}{lcc}
\toprule
\textbf{Persona} & \textbf{Cosine Sim. ($\beta$)} & \textbf{SD} \\
\midrule
P1 & 0.712 & 0.167 \\
P2 & 0.612 & 0.164 \\
P3 & 0.409 & 0.067 \\
P4 & 0.591 & 0.181 \\
P5 & 0.636 & 0.137 \\
P6 & 0.956 & 0.034 \\
\midrule
\textbf{Mean} & \textbf{0.639} & \textbf{0.219} \\
\bottomrule
\end{tabular}
\label{tab:persona_stability}
\end{table}

\subsection{Persona Representation} \label{app:narr_comp}
To interpret the qualitative substance of each of the six data-driven personas, we first identified and manually analyzed the top three narratives with the highest probability score for that archetype.
\Cref{tab:composites} details the distilled insights from that analysis.
For clarity, we present these insights as narrative composites.
These composites are not direct quotes from a single post; they are synthesized stories that weave together the key themes, events, and emotional arcs from the most representative narratives.
This technique, known as composite narrative construction~\cite{willis2019composite}, allows us to provide a single, rich, and illustrative example for each persona, complementing the concise descriptions and data visualizations in the main paper.

\input{tables/composites}

\subsection{Persona Statistics} \label{app:persona_stats}
\begin{figure}[h!]
    \centering
    \includegraphics[width=0.7\linewidth]{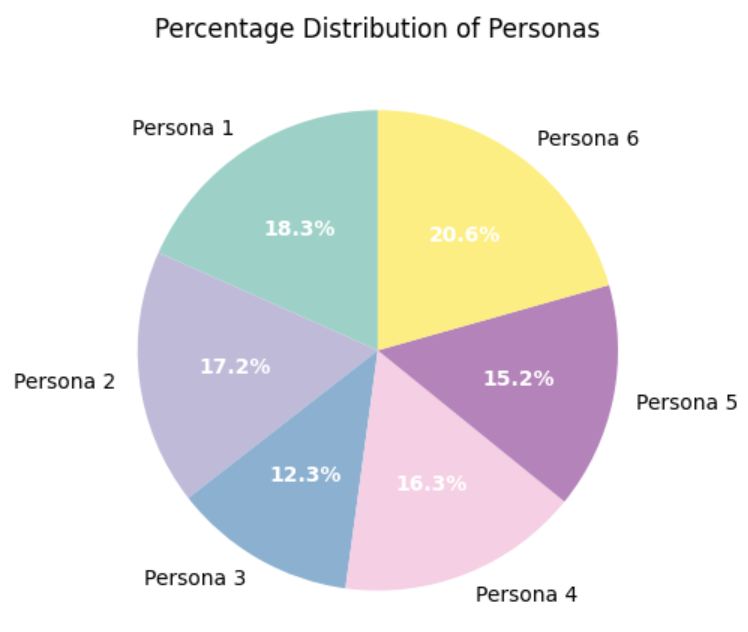}
    \caption{Overall Distribution of Personas.}
    \label{fig:persona_dist}
\end{figure}

We also provide additional descriptive statistics for the six radicalization personas discovered in our analysis.
These statistics offer further insight into their overall prevalence, their composition within individual narratives, and their evolution over time.
\Cref{fig:persona_dist} shows the average probability for each of the six personas across all analyzed profiles.

\begin{figure}[h!]
    \centering
    \includegraphics[width=0.75\linewidth]{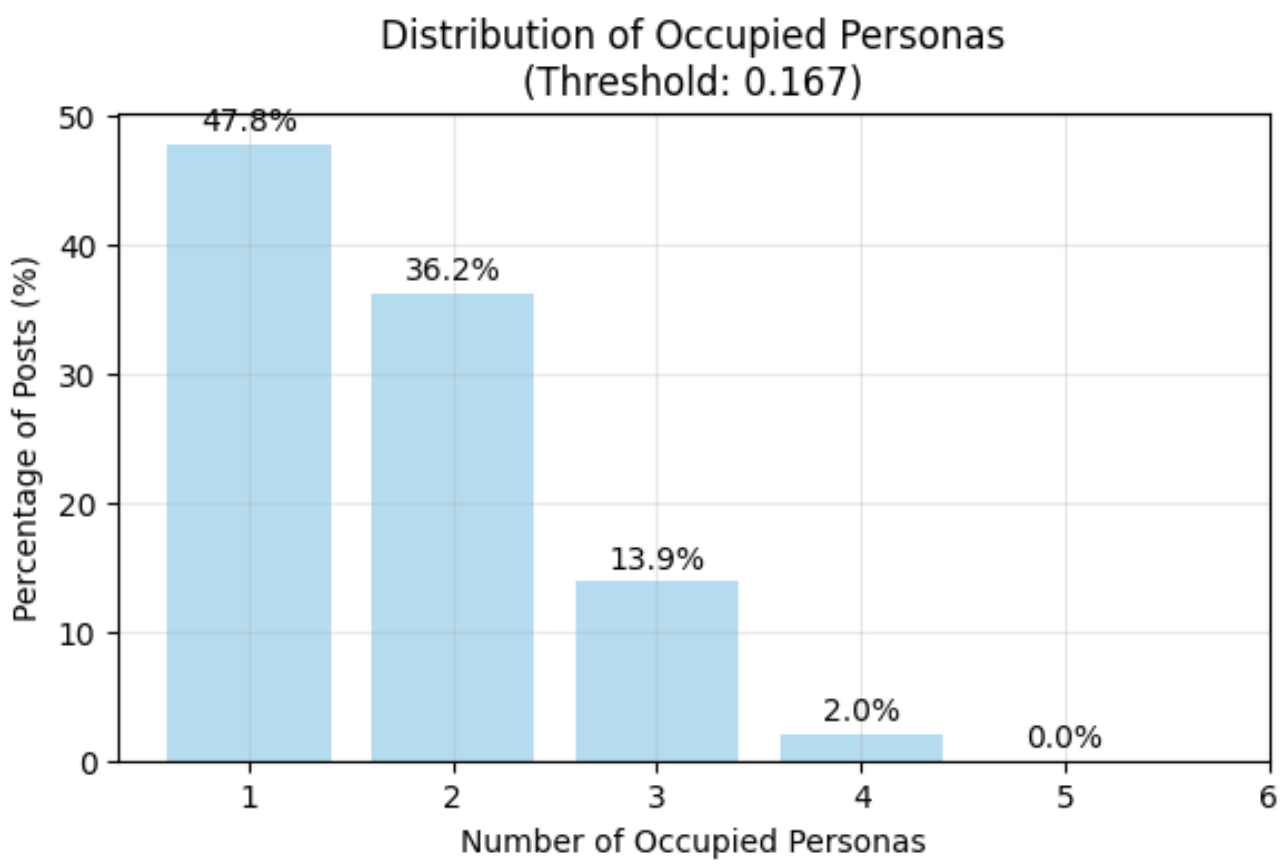}
    \caption{Distribution of Occupied Personas.}
    \label{fig:persona_occ}
\end{figure}

The distribution of occupied personas is depicted in~\Cref{fig:persona_occ}.
A profile is considered having an occupied persona $X$ if the probability assigned to persona $X$ is more than $\verb|1/6|\approx\verb|0.167|$.
Since there are six possible personas in total, a single profile may be associated with more than one persona, anywhere from one up to five at the same time.

Finally, we also illustrate the temporal dynamics of these archetypes over the time period of the dataset (2019-2024) in \Cref{fig:persona_time}:

\begin{figure}[h!]
    \centering
    \includegraphics[width=\linewidth]{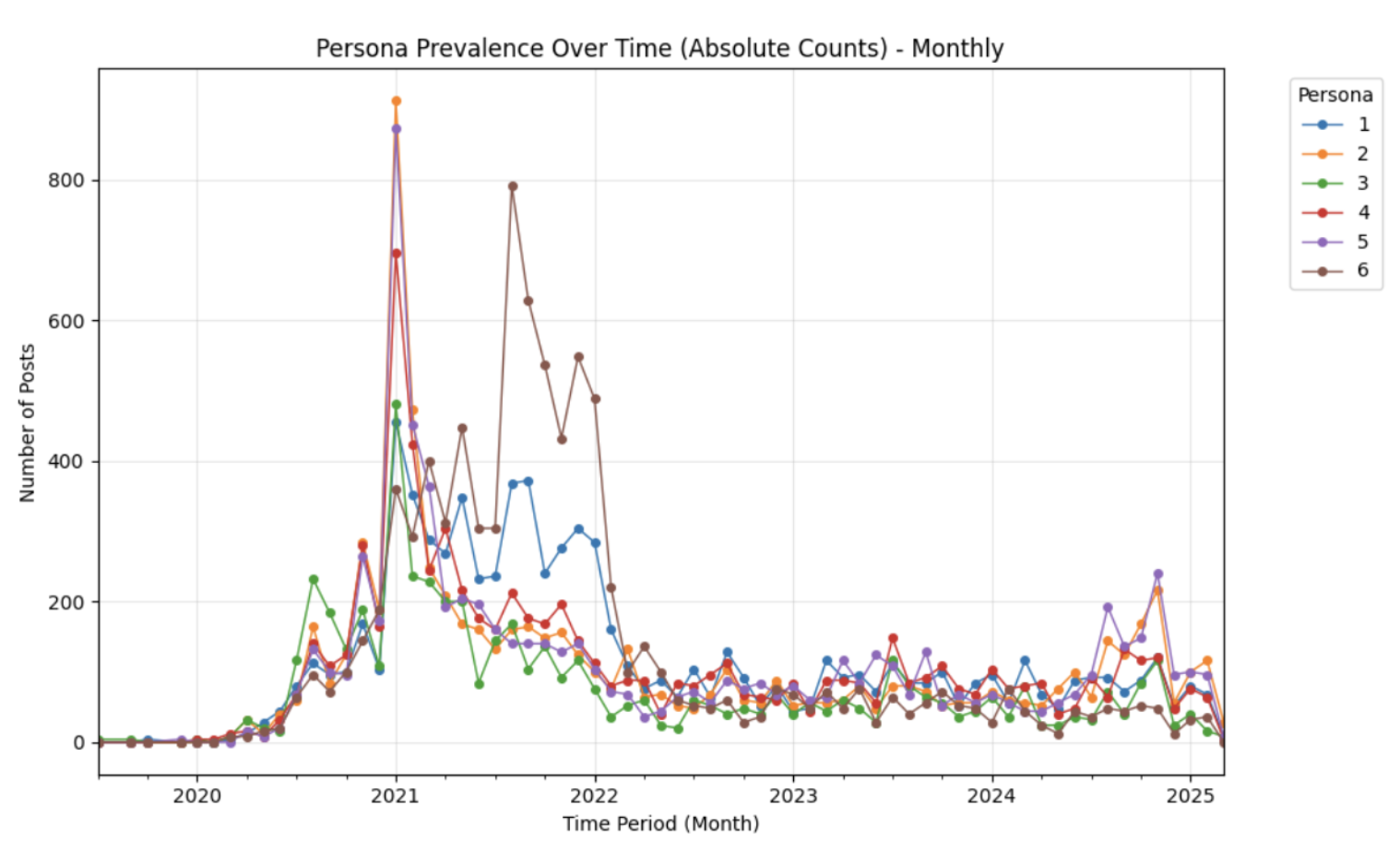}
    \caption{Evolution of Personas over Time.}
    \label{fig:persona_time}
\end{figure}

\subsection{Validation of the LLM Emotion Annotator} \label{app:emotion_ablation}
\subsubsection{Method Comparison}
To validate our choice of LLM-assisted annotation, we compare three approaches on a $100$-post sample independently annotated by two authors (annotators A and B): our GPT-based method (\texttt{gpt-4o-mini}), a lexicon-based approach (NRC Emotion Lexicon), and a fine-tuned classifier (GoEmotions; \texttt{roberta-base-go\_emotions})\footnote{\url{https://huggingface.co/SamLowe/roberta-base-go\_emotions}}.
We report Cohen's $\kappa$ between each method and each human annotator in \Cref{tab:emotion_ablation}.
GPT-based annotation substantially outperforms both baselines across all emotions and annotators, achieving mean $\kappa=0.71$ and $0.50$ against annotators A and B, respectively.
NRC and GoEmotions show near-zero agreement with human judgment (mean $\kappa \approx 0.00-0.05$), suggesting they are poorly suited to the long-form, multi-actor nature of these narratives. These results confirm that GPT-based annotation most closely approximates human judgment and is the most appropriate method for this task.
\begin{table}[h]
\centering
\scriptsize
\caption{Cohen's $\kappa$ between annotation methods and human annotators on a $100$-post sample.}
\label{tab:emotion_ablation}
\setlength{\tabcolsep}{4pt}
\begin{tabular}{llccccc}
\toprule
\textbf{Method} & \textbf{Human} & \textbf{Anger} & \textbf{Fear} & \textbf{Sadness} & \textbf{Disgust} & \textbf{Mean} \\
\midrule
GPT & A & \textbf{0.583} & \textbf{0.779} & \textbf{0.658} & \textbf{0.820} & \textbf{0.710} \\
    & B & \textbf{0.464} & \textbf{0.702} & \textbf{0.260} & \textbf{0.575} & \textbf{0.500} \\
\midrule
NRC & A & -0.020 & 0.000 & -0.016 & 0.043 & 0.002 \\
    & B & 0.019 & 0.000 & 0.138 & 0.059 & 0.054 \\
\midrule
GoEmotions & A & 0.000 & 0.077 & 0.025 & 0.000 & 0.026 \\
           & B & 0.000 & 0.090 & 0.078 & 0.000 & 0.042 \\
\bottomrule
\end{tabular}
\end{table}

\subsubsection{Annotator Agreement}
\label{app:annotator_agreement}

\revise{We provide a fuller validation of the chosen GPT-based method, reporting full confusion matrices alongside precision, recall, F1, and Cohen's $\kappa$ for each emotion against each human rater (Table~\ref{tab:llm_agreement}). The annotator achieves high raw accuracy and F1 (mean F1 = 0.91 vs. Rater A, 0.85 vs. Rater B) across all four emotions.

Error profiling shows the annotator is high-recall (92--100\%) and its errors are dominated by false positives (24 of 27 disagreement cases), meaning it rarely misses a genuine emotional expression but sometimes over-attributes one; Appendix~\ref{app:error_analysis} provides a qualitative taxonomy of these cases.

We note one apparent inconsistency: $\kappa$ for sadness is low against
Rater B (0.260) despite high raw agreement (90\%) and F1 (0.946). We
attribute this to the prevalence paradox \citep{feinstein1990high}---
because sadness is expressed in 96\% of the validation sample, chance
agreement is extremely high ($P_e \approx 0.865$), which mechanically
depresses $\kappa$ even when raw agreement and classifier utility remain
strong. We therefore report raw accuracy and F1 alongside $\kappa$
throughout, as we take the former to better reflect classifier utility in
this skewed regime.}

\begin{table}[h]
\centering
\scriptsize
\caption{Full agreement metrics between the LLM annotator and each of two human raters, per emotion, on the 100-post validation sample.}
\label{tab:llm_agreement}
\begin{tabular}{llccccc}
\toprule
\textbf{Human} & \textbf{Emotion} & \textbf{Prec.} & \textbf{Rec.} & \textbf{F1} & \textbf{Acc.} & \textbf{$\kappa$} \\
\midrule
A & Anger   & 0.700 & 1.000 & 0.824 & 79.0\% & 0.583 \\
 & Fear    & 0.857 & 0.941 & 0.897 & 89.0\% & 0.779 \\
 & Sadness & 0.980 & 1.000 & 0.990 & 98.0\% & 0.658 \\
 & Disgust & 0.906 & 0.923 & 0.914 & 91.0\% & 0.820 \\
\midrule
B & Anger   & 0.657 & 0.939 & 0.773 & 73.0\% & 0.464 \\
 & Fear    & 0.786 & 0.936 & 0.854 & 85.0\% & 0.702 \\
 & Sadness & 0.898 & 1.000 & 0.946 & 90.0\% & 0.260 \\
 & Disgust & 0.868 & 0.767 & 0.814 & 79.0\% & 0.575 \\
\bottomrule
\end{tabular}
\end{table}

\subsubsection{Qualitative Error Analysis}
\label{app:error_analysis}

\revise{We conducted a qualitative analysis of the 27 cases where the LLM disagreed with both human annotators (who were themselves in agreement), identifying four recurring error categories:

\begin{itemize}
    \item \textbf{Source misattribution (9):} the model assigns an emotion expressed by a quoted third party (e.g., a family member's own words within the post) to the narrator, rather than tracking the narrator's actual stance toward that content. For instance, one post quotes a father's angry reply accusing his daughter of unfairly calling him racist; the model flagged the post as expressing disgust, even though the disgust belonged to the quoted father, while the narrator's own reaction throughout the post was one of sadness at the relationship's breakdown.
    \item \textbf{Lexical bleed from topic content (8):} emotion-adjacent vocabulary (fear/threat language, violent content) triggers a positive prediction independent of the narrator's actual emotional register, which is often closer to grief, exhaustion, or contempt. For example, a post describing a mother who wished serious illness on people who voted differently, and who threatened family members over politics, was flagged for fear despite the narrator's own tone reading as weary contempt rather than fear.
    \item \textbf{Register/rhetorical confusion (6):} sarcasm, dismissive idiom, and rhetorical questions carry surface intensity markers that are misread as the literal emotion, when the narrator's stated or contextual affect is different (and sometimes named explicitly elsewhere in the same post). For example, a narrator closed a post with a sarcastic line blaming several public figures and institutions for estranging them from a family member; the model read this as anger, even though the narrator explicitly named their emotion as sadness a few lines earlier.
    \item \textbf{Implicit/under-lexicalized emotion underweighted (4):} when an emotion is conveyed through consequence, somatic description, or moral disapproval rather than explicit affect vocabulary, the model tends to miss it, suggesting a reliance on explicit lexical cues over pragmatic inference. For instance, a healthcare worker described repeatedly having to intubate COVID patients and notify families of deaths while trying to protect her infant son from unvaccinated relatives; despite the clear underlying fear, the model did not detect this emotion, likely because it was conveyed through professional and clinical detail rather than explicit fear-related wording.

\end{itemize}
}

\subsection{Regex for Control Variable Extraction} \label{app:regex}
The high-precision regular expression patterns used to create the ground-truth dataset for validating our LLM-based control variable extraction are inspired by prior work on demographic inference from social media text~\cite{tigunova2020reddust}, refining their patterns based on common phrasing observed in our dataset.
\Cref{tab:regex_patterns} outlines the general syntactic patterns we search for, while \Cref{tab:regex_indicators} lists the specific keywords and phrases used as indicators within those patterns. 

\begin{figure}
    \centering
    \includegraphics[width=0.75\linewidth]{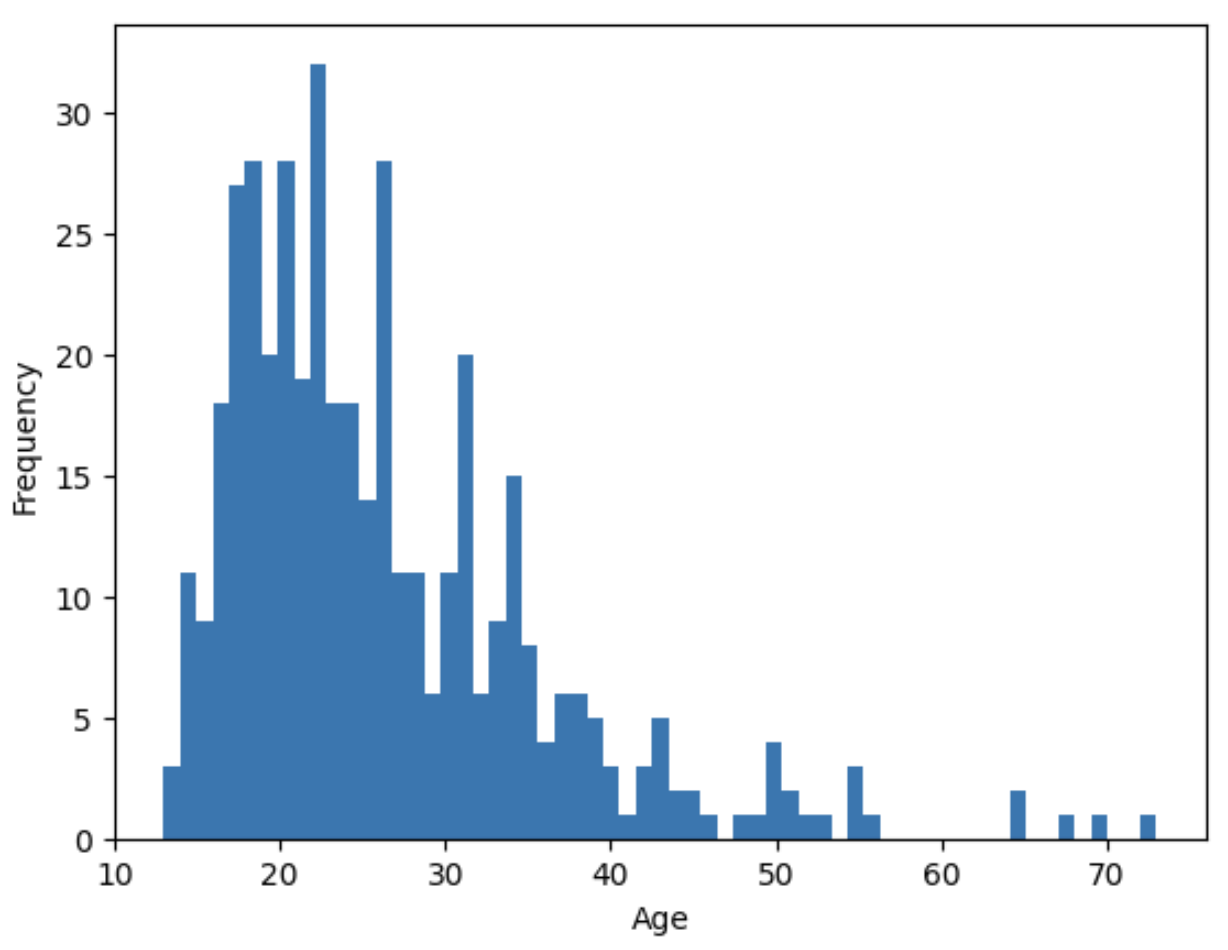}
    \caption{Distribution of ages of authors.}
    \label{fig:age_freq}
\end{figure}

After identifying these variables, \emph{gender} is coded as a binary variable (male or female), and \emph{age group} is grouped by adolescence (13-20) and adult (21+) for the demonstrated usage and performance in social media analysis~\cite{cesare2017well} and the distribution of user's age shown in \Cref{fig:age_freq}

\begin{table}[h!]
\scriptsize
\centering
\caption{Syntactic patterns for labeling narrator age and gender. Placeholders like `<...>` are filled with indicators from \Cref{tab:regex_indicators}.}
\label{tab:regex_patterns}
\begin{tabular}{ll}
\toprule
\textbf{Attribute} & \textbf{Pattern(s)} \\
\midrule
\textbf{gender} & (I am | I'm) a? `<gender indicator>` \\
\addlinespace
\textbf{age} & (i) (I am | I'm) `<number>` years old \\
& (ii) I'm a `<number>`-year-old `<gender indicator>` \\
& (iii) I am `<number>` followed by punctuation/conjunction \\
& (iv) Compact Format: `(<number><gender indicator>)` \\
\bottomrule
\end{tabular}
\end{table}

\begin{table}[h!]
\scriptsize
\centering
\caption{Word and phrase indicators used within the patterns from \Cref{tab:regex_patterns}.}
\label{tab:regex_indicators}
\begin{tabular}{lll}
\toprule
\textbf{Attribute} & \textbf{Value} & \textbf{Word/Phrase Indicators} \\
\midrule
\textbf{gender} & female & woman, female, girl, lady, wife, mother, sister \\
& male & man, male, boy, guy, husband, father, brother \\
\bottomrule
\end{tabular}
\end{table}

\subsection{Full Emotion Regression Coefficient Table}
The full regression coefficients for RQ2 are shown in \Cref{tab:reg_coeff}.
The categorical control variables are coded using effect coding, where we interpret their effect from the overall mean, given there is no natural baseline for choosing a reference category (e.g, among different family relationships). 

\begin{table*}[ht!]
\sisetup{table-number-alignment=right,table-format=1.3}
\centering
\caption{Regression Results (Odds Ratios and p-values, rounded to 3 decimals). Significance levels: *** p $<$ 0.001, ** p $<$ 0.01, * p $<$ 0.05.}
\label{tab:reg_coeff}
\scriptsize
\begin{tabular}{lSSSSSSSS}
\toprule
 & \text{OR (anger)} & \text{p (anger)} & \text{OR (disgust)} & \text{p (disgust)} & \text{OR (fear)} & \text{p (fear)} & \text{OR (sadness)} & \text{p (sadness)} \\
\midrule
Intercept & 2.305*** & 0.000 & 1.509*** & 0.000 & 1.656*** & 0.000 & 7.933*** & 0.000 \\
C(relationship\_cat)[extended\_family] & 1.038 & 0.075 & 1.231 & 0.058 & 0.988 & 0.912 & 0.826 & 0.255 \\
C(relationship\_cat)[friend] & 0.615*** & 0.000 & 0.995 & 0.962 & 0.665*** & 0.000 & 0.949 & 0.075 \\
C(relationship\_cat)[parent] & 1.458*** & 0.000 & 1.090 & 0.092 & 1.151** & 0.006 & 1.331** & 0.001 \\
C(relationship\_cat)[partner] & 0.733*** & 0.000 & 0.664*** & 0.000 & 1.355*** & 0.000 & 0.691*** & 0.000 \\
C(age\_group)[adolescence] & 0.878*** & 0.001 & 0.922* & 0.016 & 1.174*** & 0.000 & 0.776*** & 0.000 \\
C(gender)[f] & 0.918* & 0.038 & 0.898** & 0.000 & 0.980 & 0.059 & 1.098 & 0.113 \\
balance\_1 & 0.927** & 0.041 & 0.870*** & 0.000 & 1.045 & 0.069 & 0.928 & 0.067 \\
balance\_2 & 1.036 & 0.230 & 0.878*** & 0.000 & 1.087** & 0.023 & 1.109* & 0.031 \\
balance\_3 & 1.024 & 0.343 & 0.973 & 0.214 & 0.897*** & 0.000 & 1.108* & 0.012 \\
balance\_4 & 1.108** & 0.050 & 1.086*** & 0.000 & 0.936*** & 0.009 & 1.025 & 0.057 \\
balance\_5 & 0.999 & 0.977 & 0.898*** & 0.000 & 0.986 & 0.057 & 0.989 & 0.808 \\
\bottomrule
\end{tabular}
\end{table*}

%% file: tables/phase3-topics.tex
\begin{table*}[ht!]
\centering
\caption{Hierarchical structure of the Phase 3 (\emph{Post-Radicalization}) topic consolidation. The rightmost column shows examples of the original, fine-grained topics that were grouped into the Sub-Categories and Categories (shown in the main Sankey) on the left.}
\label{tab:phase3_topics}
\scriptsize
\renewcommand{\arraystretch}{1.05}
\begin{tabular}{p{2cm} l l}
\toprule
\multirow{39}{*}{\parbox{2cm}{Behavioral Transformation (External)}}
 & \multirow{4}{*}{Social Deterioration and Relationship Strain}       & Pushback Against Boundaries                                        \\
                                                       &                                                                     & Communication Tension and Breakdown                                \\
                                                       &                                                                     & Social Isolation and Limited Friendships                           \\
                                                       &                                                                     & Family Estrangement and Disconnection                              \\ \cline{2-3} 
                                                       & \multirow{5}{*}{Interpersonal Harm and Moral Disregard}             & Unfaithful Apology                                                 \\
                                                       &                                                                     & Increased Aggression and Hostility Toward Others                   \\
                                                       &                                                                     & Loss of Compassion and Empathy                                     \\
                                                       &                                                                     & Framing Others as Indoctrinated                                    \\
                                                       &                                                                     & Emotional and Physical Abuse                                       \\ \cline{2-3} 
                                                       & \multirow{5}{*}{Alternative Media Use and Information Exposure}     & Distrust in Mainstream Media                                       \\
                                                       &                                                                     & Joining and Using Alternative Social Media Platforms (Parler, Gab) \\
                                                       &                                                                     & Using Telegram for QAnon Information                               \\
                                                       &                                                                     & Excessive Screen Time and Online Content Consumption               \\
                                                       &                                                                     & Disseminating QAnon Content and Persuading Others                  \\ \cline{2-3} 
                                                       & \multirow{5}{*}{Lifestyle Changes Driven by Conspiratorial Beliefs} & Increased Gun Acquisition                                          \\
                                                       &                                                                     & Precious Metals Investment                                         \\
                                                       &                                                                     & Prepping and Hoarding Supplies                                     \\
                                                       &                                                                     & Homeschooling Children                                             \\
                                                       &                                                                     & Employment Instability and Career Disruption                       \\ \cline{2-3} 
                                                       & \multirow{5}{*}{Psychological and Emotional Shifts}                 & Manifestations of Intense Hatred                                   \\
                                                       &                                                                     & Collective Narcissism                                              \\
                                                       &                                                                     & Inflated Self-Perception as Intellectual Superiority               \\
                                                       &                                                                     & Willful Ignorance                                                  \\
                                                       &                                                                     & Personality Changes                                                \\ \cline{2-3} 
                                                       & \multirow{4}{*}{Radical Ideological Alignment and Intolerance}      & Racism Attitudes                                                   \\
                                                       &                                                                     & Negative Views on LGBTQ+ Communities                               \\
                                                       &                                                                     & Fascism Accusations and Rhetoric                                   \\
                                                       &                                                                     & Deflecting to Race-Based Movements                                 \\ \cline{2-3} 
                                                       & \multirow{5}{*}{Public Health Mistrust and Resistance}              & Vaccine Refusal and Hesitancy                                      \\
                                                       &                                                                     & Denying COVID-19 Reality                                           \\
                                                       &                                                                     & Resistance to Mask-Wearing                                         \\
                                                       &                                                                     & Resistance to Therapy and Counseling                               \\
                                                       &                                                                     & Distrust in Medical Professionals                                  \\ \cline{2-3} 
                                                       & \multirow{6}{*}{Epistemic Rigidity and Distrust in Expertise}       & Rejecting Verified Sources and Clinging to Unverified Information  \\
                                                       &                                                                     & Distrust in Science                                                \\
                                                       &                                                                     & Using "Science" to Justify Conspiracy Beliefs                      \\
                                                       &                                                                     & Insistence on Personal Research as Evidence                        \\
                                                       &                                                                     & Uncritical Acceptance of Beliefs and Theories                      \\
                                                       &                                                                     & Lack of Critical Thinking                                          \\ \midrule
\multirow{37}{*}{Beliefs (Internal)}                   & \multirow{4}{*}{Child Harm and Elite Abuse Narratives}              & Child Sex Trafficking and Satanic Sacrifice                        \\
                                                       &                                                                     & Pizzagate                                                          \\
                                                       &                                                                     & Epstein-related Beliefs                                            \\
                                                       &                                                                     & Disney "Hidden Imagery"                                            \\ \cline{2-3} 
                                                       & \multirow{7}{*}{Apocalyptic and Eschatological Narratives}          & The Cabal and "The Storm"                                          \\
                                                       &                                                                     & Apocalyptic and Predictive Programming                             \\
                                                       &                                                                     & "10 Days of Darkness"                                              \\
                                                       &                                                                     & "Trust the Plan"                                                   \\
                                                       &                                                                     & Lizard People                                                      \\
                                                       &                                                                     & Religious End Times                                                \\
                                                       &                                                                     & Alien                                                              \\ \cline{2-3} 
                                                       & \multirow{6}{*}{Pro-Trump}                                          & Trump's Return and January 6th Insurrection                        \\
                                                       &                                                                     & Trump as a Religious Figure                                        \\
                                                       &                                                                     & MAGA                                                               \\
                                                       &                                                                     & Biden Presidency Denial                                            \\
                                                       &                                                                     & Negative Reactions to Kamala Harris and Political Opposition       \\
                                                       &                                                                     & JFK Jr. Being Alive                                                \\ \cline{2-3} 
                                                       & \multirow{3}{*}{Anti-Government and Institutional Distrust}         & Deep State                                                         \\
                                                       &                                                                     & New World Order                                                    \\
                                                       &                                                                     & Election Fraud                                                     \\ \cline{2-3} 
                                                       & \multirow{4}{*}{Medical Mistrust}                                   & Drinking Bleach as a Cure                                          \\
                                                       &                                                                     & Ivermectin Use in Treating Illness                                 \\
                                                       &                                                                     & Pandemic Misinformation                                            \\
                                                       &                                                                     & Dr. Fauci Creating COVID                                           \\ \cline{2-3} 
                                                       & \multirow{4}{*}{Science and Technology Mistrust}                    & Climate Change is not Real                                         \\
                                                       &                                                                     & 5G Being Dangerous                                                 \\
                                                       &                                                                     & Fake Moon Landing                                                  \\
                                                       &                                                                     & Flat Earth                                                         \\ \cline{2-3} 
                                                       & \multirow{3}{*}{War-Inducing Rhetoric}                              & Antisemitic Beliefs and Holocaust Denial                           \\
                                                       &                                                                     & Russia-Ukraine War Propaganda                                      \\
                                                       &                                                                     & Invoking War, Freedom, and the Constitution                        \\ \cline{2-3} 
                                                       & \multirow{2}{*}{Defamatory Conspiracy Targeting Public Figures}     & Bill Gates                                                         \\
                                                       &                                                                     & Michelle Obama Being a Man                                         \\ \cline{2-3} 
                                                       & \multirow{2}{*}{Policy Defiance}                                    & Tax Evasion                                                        \\
                                                       &                                                                     & Abortion Opposition                                                \\ \cline{2-3} 
                                                       & \multirow{2}{*}{Anti-Communism and Geopolitical Distrust}           & China and Global Influence Concerns                                \\
                                                       &                                                                     & Criticism towards Communism and Socialism                          \\
                                                       \bottomrule

\end{tabular}
\end{table*}

%% file: tables/composites.tex
\begin{table*}[!ht]
\renewcommand{\arraystretch}{0.7}
\scriptsize
\begin{center}
\caption{The narrative composites representing six radicalization personas.}
\label{tab:composites}
\begin{tabularx}{\textwidth}{l X}
\toprule
\textbf{Persona} & \textbf{Narrative Composite} \\
\midrule
The Health-Triggered Conspiracy Theorist & He used to be such a bright, intelligent, and sensible person. We were close friends since grade school, and he had a promising career as an engineer. But over the last couple of years, things took a dark turn. It started with some underlying health issues and escalated as he began mixing substances he was buying online.
He started having these terrifying psychotic episodes, ranting about Q-adjacent theories and believing he was being watched. It got so bad his roommates had to call the police to have him hospitalized. He lied to me about it, claiming the doctors were performing experiments on him. He refuses to get professional help because he’s convinced all doctors are evil and will just ``inject him with shit.'' He eventually lost his job, and his spouse and I are watching him deteriorate. Now he spends his days in front of the computer, convinced the online community he interacts with knows more than anyone, even ranting at his wife---who is a nurse---about miracle cures he can't even pronounce. He used to be my best friend, and now he's just a shell of a human.
 \\ \midrule
The Political Extremist        & My relationship with a close family member has been completely destroyed by their political identity. They don't just support Trump; they worship him like a deity. Their entire worldview is based on hatred, fear, and a perceived sense of victimization, a constant us versus them mentality that echoes the darkest moments in history. This isn't hyperbole; the rhetoric is genuinely terrifying.
They are trapped in a right-wing media bubble. I've tried to reason with them, carefully vetting my own sources to present non-partisan facts, but anything that contradicts their narrative is dismissed as ``MSM Propaganda.'' Their radicalization has had severe real-world consequences: they lost their job as a teacher over their social media posts, were present at the January 6th riot, and have alienated everyone who tries to get close to them. We even attempted an intervention to show them how these beliefs were ruining their life, but it ended with them getting angry and trying to start a fight. I've had to end the relationship because I can't be close to someone whose values have become so full of hate. \\ \midrule
The Social Media Spiral      & My friend is very dear to me, but over the last few years, she's been consumed by conspiracy-minded YouTubers and alt-right talking points. She's completely Q-adjacent, believing everything they tell her to fear and be angry about. She thinks she's thinking for herself by using alternative media sources, but she's just being misled by paranoid video bloggers.
Her entire reality is now shaped by these online rabbit holes. She's obsessed with wild theories about public figures being involved in child harm and elite abuse, things that started on fringe message boards like 8chan. It’s a gigantic, ridiculous lie, but she takes it as gospel. I feel terrible for the people she and thousands of others baselessly accuse online of running satanic cults and using children's blood for anti-aging cream. When she visits, I want to see her, but I dread hearing about the latest thing some influencer said. It's exhausting, like they've co-opted the idea of breaking free from mental control only to submit to another, even uglier, form of it. \\ \midrule
The Religious Apocalypticist       & My wife used to be my partner, but now she's lost in a world where conspiracy theories and religious prophecy are one and the same. It started years ago with commentators like Alex Jones and spiraled into QAnon around 2018. She now believes in everything: the deep state, lizard people, adrenochrome, all of it. For her, and many in her conservative Christian circles, QAnon isn't just a conspiracy; it's a confirmation that the end times are here.
Every major event is filtered through this lens. The COVID vaccine was the ``mark of the beast,'' a tool for population control created by Bill Gates. She was convinced something huge would happen on January 6th to keep Trump in power. While she no longer believes ``the storm is coming'' after so many failed predictions, the underlying conspiratorial worldview remains. As a fellow Christian, I know that no one knows the day or the hour, but for her, these theories have become a self-reinforcing prophecy. She has cut off family and friends who don't agree, and I feel like I'm living with a stranger.
 \\ \midrule
The Conservative Identity Protector & My parents used to be people I considered good and loving. But since 2016, my mother has transformed into someone I don't recognize, consumed by a need to protect her conservative identity from a world she sees as a threat. She is a Black woman who has become obsessed with right-wing talk radio and fringe news, and her mind is deteriorating as a result. She is terrified that our kids are being ``transed,'' sees an ``invasion at the border,'' and is constantly making snide remarks about anyone who doesn't share her views.
Our family is suffering. Her constant political ramblings and racist, bigoted comments have created a screaming hell in our own house. As her gay son, happily married and trying to start a family, it's devastating to know she votes for people who want to strip away my rights. After the last election, she texted me a series of cheerful emojis, asking how I was ``holding up''---not out of concern, but to bait me into an argument and gloat. She wants to ``own the libs'' more than she wants a relationship with her own son. It's exhausting and I am so done trying to talk sense into her.
 \\ \midrule
The Pandemic-Triggered Skeptic             & My mom has always leaned towards alternative medicine, but the COVID-19 pandemic sent her over the edge. We managed to get through the first few years by tiptoeing around the subject, but now it's unavoidable. It started with soft questions about how many vaccines I'd gotten, and now it's full-blown propaganda. She's convinced the vaccines are a form of mass genocide designed to alter our DNA and make us infertile.
She calls herself a truth seeker and spends her time on platforms like Telegram, deep in rabbit holes about satanic abuse and globalist plots. She rejects the term ``conspiracy theory'' and proudly states she'd rather lose her job than get vaccinated. I made the mistake of checking out a YouTube channel she recommended, and it was pure fear-mongering and right-wing propaganda. We used to be thick as thieves, but now I feel like I'm mourning the person she once was. Her pre-existing skepticism of mainstream medicine was the gateway, and the pandemic pushed her through it into a much darker world.
 \\
\bottomrule
\end{tabularx}
\end{center}
\end{table*}